\pdfoutput=1

\RequirePackage[l2tabu, orthodox]{nag}
\documentclass[12pt]{article}

\usepackage[T1]{fontenc}
\usepackage{microtype}

\usepackage{amsmath}
\usepackage{amssymb}
\usepackage{amsthm}
\usepackage{mathtools}
\usepackage{thmtools}
\usepackage{mathrsfs}

\usepackage{libertine}      

\usepackage{titlesec}

\titleformat*{\section}{\Large\bfseries}

\usepackage[usenames,dvipsnames]{xcolor}
\definecolor{shadecolor}{gray}{0.9}

\usepackage[english]{babel}
\usepackage[parfill]{parskip}
\usepackage{afterpage}
\usepackage{framed}
\usepackage{nicefrac}

{\endMakeFramed}

\DeclareRobustCommand{\parhead}[1]{\textbf{#1}~}

\newcounter{parcount}

\usepackage{graphicx}
\usepackage[labelfont=bf]{caption}
\usepackage[format=hang]{subcaption}
\usepackage{wrapfig}

\usepackage{booktabs}
\usepackage{multirow}
\usepackage{longtable}

\usepackage[sort]{natbib}
\usepackage{bibunits}

\usepackage[algoruled]{algorithm2e}
\usepackage{listings}
\usepackage{fancyvrb}
\fvset{fontsize=\normalsize}

\usepackage{hyperref}
\hypersetup{colorlinks,linktoc=all}
\hypersetup{citecolor=Violet}
\hypersetup{linkcolor=black}
\hypersetup{urlcolor=MidnightBlue}

\usepackage[nameinlink]{cleveref}

\usepackage[acronym,nowarn]{glossaries}

\usepackage{listings}
\usepackage{lstbayes}
\lstdefinestyle{mystyle}{
    commentstyle=\color{OliveGreen},
    numberstyle=\tiny\color{black!60},
    stringstyle=\color{BrickRed},
    basicstyle=\ttfamily\scriptsize,
    breakatwhitespace=false,
    breaklines=true,
    captionpos=b,
    keepspaces=true,
    numbers=none,
    numbersep=5pt,
    showspaces=false,
    showstringspaces=false,
    showtabs=false,
    tabsize=2
}
\usepackage{enumitem}

\crefname{lemma}{lemma}{lemmas}
\Crefname{lemma}{Lemma}{Lemmas}
\crefname{thm}{theorem}{theorems}
\Crefname{thm}{Theorem}{Theorems}
\crefname{prop}{proposition}{propositions}
\Crefname{prop}{Proposition}{Propositions}

\newtheorem{thm}{Theorem} 
\newtheorem{defn}[thm]{Definition} 
\newtheorem{prop}[thm]{Proposition}
\newtheorem{lemma}[thm]{Lemma}

\setacronymstyle{long-sc-short}
\newacronym{KL}{kl}{Kullback-Leibler}

\newacronym{POPELBO}{pop-elbo}{\emph{population evidence lower bound}}
\newacronym{PROELBO}{pro-elbo}{\emph{profile evidence lower bound}}

\newacronym{SVI}{svi}{stochastic variational inference}
\newacronym{VI}{vi}{variational inference}

\newacronym{ADVI}{advi}{automatic differentiation variational inference}

\newacronym{LDA}{lda}{latent Dirichlet allocation}

\newacronym{SMC}{smc}{Sequential Monte Carlo}
\newacronym{VB}{vb}{variational Bayes}

\newacronym{TDVI}{tdvi}{transdimensional variational inference}
\newacronym{STDVI}{stdvi}{sequential transdimensional variational inference}
\newacronym{MCMC}{mcmc}{Markov chain Monte Carlo}
\newacronym{RJMCMC}{rjmcmc}{reversible jump Markov chain Monte Carlo}
\newacronym{TDMCMC}{tdmcmc}{transdimensional Markov chain Monte Carlo}

\newacronym{SLDS}{slds}{switching linear dynamical system}
\newacronym{HDP-SLDS}{hdp-slds}{hierarchical Dirichlet process switching linear dynamical system}

\newacronym{MLM}{mlm}{masked language model}
\newacronym{CBOW}{cbow}{continuous bag of words}
\newacronym{MoE}{moe}{mixture-of-experts}
\newacronym{SGNS}{sgns}{skip-gram with negative sampling}
\newacronym{LLM}{llm}{large language model}
\newacronym{AWE}{awe}{attention word embedding}
\newacronym{OOD}{ood}{out-of-distribution}

\usepackage[margin=1in]{geometry}
\usepackage{tcolorbox}
\usepackage{adjustbox}

\usepackage[defaultlines=3,all]{nowidow}

\usepackage{libertine}  
\usepackage{dsfont}

\usepackage{caption}
\usepackage{subcaption}
\usepackage{wrapfig}

\DeclareCaptionFormat{empty}{}

\title{Bidirectional Attention as a \\ Mixture of Continuous Word Experts}

\author{
  Kevin Christian Wibisono\\
        University of Michigan, Statistics\\
  kwib@umich.edu\\
  \and
  Yixin Wang\\
          University of Michigan, Statistics\\
  yixinw@umich.edu\\
  }

\date{\today}

\begin{document}
\maketitle


\begin{abstract}
Bidirectional attention---composed of self-attention with positional encodings and the
\gls{MLM} objective---has emerged as a key component of modern
\glspl{LLM}. Despite its empirical success, few studies have examined
its statistical underpinnings: What statistical model is bidirectional
attention implicitly fitting? What sets it apart from its
non-attention predecessors? We explore these questions in this paper.
The key observation is that fitting a single-layer single-head
bidirectional attention, upon reparameterization, is equivalent to
fitting a \gls{CBOW} model with \gls{MoE} weights. Further,
bidirectional attention with multiple heads and multiple layers is
equivalent to stacked \glspl{MoE} and a mixture of
\glspl{MoE}, respectively. This statistical viewpoint reveals the
distinct use of \gls{MoE} in bidirectional attention, which aligns
with its practical effectiveness in handling heterogeneous data. It
also suggests an immediate extension to categorical tabular data, if
we view each word location in a sentence as a tabular feature. Across
empirical studies, we find that this extension outperforms existing
tabular extensions of transformers in \gls{OOD} generalization.
Finally, this statistical perspective of bidirectional attention
enables us to theoretically characterize when linear word analogies
are present in its word embeddings. These analyses show that
bidirectional attention can require much stronger assumptions to
exhibit linear word analogies than its non-attention predecessors.\footnote{Software that
  replicates the empirical studies can be found at \url{https://github.com/yixinw-lab/attention-uai}.}
\end{abstract}


\glsresetall

\section{Introduction}

\label{sec:intro}

First introduced by \citet{vaswani2017attention}, bidirectional attention represents a departure from the
traditional recurrent or convolutional neural networks in language
modeling. This architecture has since become the backbone of many
large language models, including BERT~\citep{devlin2018bert},
RoBERTa~\citep{liu2019roberta}, and GPT-2~\citep{radford2019language},
all of which have achieved exceptional performance in natural language
processing benchmarks.

At the heart of bidirectional attention lies the self-attention mechanism: it creates a holistic representation of each sentence by capturing pairwise relationships between tokens in the sentence. Equally important are positional encodings, supplying word ordering information that allows bidirectional attention to move beyond bag-of-words. Finally, bidirectional
attention employs the \gls{MLM} objective, a self-supervised learning objective for unlabelled text data which minimizes the model's cross-entropy loss for predicting randomly masked words within each sentence.

Despite the empirical success of attention-based language models, few works have examined their statistical underpinnings: What statistical models are these attention-based models implicitly fitting? What sets these models apart from their non-attention predecessors like
\gls{CBOW}~\citep{mikolov2013distributed}? How does the use of the self-attention mechanism contribute to these models' empirical success? We
explore these questions in this work.

\parhead{Main ideas and contributions.} In this paper, we conduct a theoretical investigation into bidirectional attention. The key observation is that upon reparameterization, fitting a single-head and single-layer bidirectional attention is equivalent to fitting
\gls{CBOW} with \gls{MoE} weights~\citep{jacobs1991adaptive}.
Moreover, bidirectional attention with multiple heads and multiple
layers is equivalent to stacked \glspl{MoE} and a mixture of \glspl{MoE}, respectively. These analyses reveal the distinct use of \gls{MoE} in bidirectional attention as compared with its
non-attention predecessors. In particular, they partially explain the practical
effectiveness of bidirectional attention in capturing heterogeneous natural language patterns~\citep{devlin2018bert,liu2019roberta}.

This statistical interpretation of bidirectional attention suggests an immediate extension to modeling (categorical) tabular data: one can view each word position in a sentence as a
tabular feature, and each word as the value that the feature takes. Across empirical studies, we find that this tabular extension improves \gls{OOD} generalization compared with existing
tabular data algorithms or tabular extensions of attention. Moreover, this extension may facilitate the integration of heterogeneous data sets with partially overlapping features: the
learned feature encodings (akin to positional encodings in the original attention module) bring all features into the same embedding
space.

Finally, the connection between bidirectional attention and \gls{CBOW}+\gls{MoE} empowers us to theoretically characterize when linear
word analogies (e.g. $\mathrm{king} - \mathrm{man} + \mathrm{woman} \approx
\mathrm{queen}$) can be present in its word embeddings. We draw on a classical finding in \citet{levy2014neural}: the similarity between
two tokens from word2vec embeddings is equal to their pointwise mutual
information~\citep{church1990word}, provided that the embeddings have
sufficient dimensionality and the models are trained using the \gls{SGNS} objective. This result enables us to analyze the embeddings
of bidirectional attention through its connection to \gls{CBOW}. Adopting the paraphrasing argument of \citet{allen2019analogies} for
\gls{SGNS}, we characterize the conditions under which both \gls{CBOW} and attention-based embeddings exhibit linear word analogies. We show
that bidirectional attention can require much stronger conditions than its non-attention predecessors. These results partially explain the empirical observations that bidirectional attention may not always achieve meaningful improvements over classical word embeddings in capturing abstract and complex relationships~\citep{ushio2021bert}.


\parhead{Related work.} Our work draws on three themes around
attention-based models.

The first is a body of work on the \textit{theoretical foundations of
attention-based models}. \citet{elhage2021mathematical} analyzed how
the different components of decoder-only attention-based architectures
relate to each other. \citet{edelman2022inductive} provided a rigorous
justification of the ability of attention-based architectures to
represent sparse functions. \citet{tsai2019transformer} viewed
attention through the perspective of kernels. \citet{peng2020a}
established a connection between the use of multiple heads in
transformers and \gls{MoE}. \citet{li2023transformers} showed that the
embedding and self-attention layers in a transformer architecture are
capable of capturing topic structures. 

More recently, numerous studies have been devoted to unraveling the reasons behind the exceptional performance of attention-based models, such as transformers, at in-context learning from various viewpoints. For instance, ~\citet{akyurek2022learning, oswald2023transformers,dai2023why,bai2023transformers,zhang2023trained}, and \citet{ahn2023transformers} attributed this intriguing phenomenon to transformers' capability to implement gradient descent. Alternatively, studies by \citet{xie2021an}, \citet{ahuja2023in}, and \citet{wang2023large} approached the explanation from a Bayesian perspective. In contrast to these works, we provide a statistical interpretation of the bidirectional attention objective,
showing that fitting a single-layer single-head attention-based architecture is equivalent to fitting a \gls{CBOW} model with \gls{MoE} weights; this statistical interpretation provides a theoretical basis for the empirical effectiveness of bidirectional attention in handling heterogeneous data~\citep{devlin2018bert,liu2019roberta}.

The second theme is the \textit{extension of attention-based models to
tabular data}. One prominent work along this line is TabTransformer
\citep{huang2020tab}, which utilizes a concatenation of token embeddings and unique feature
identifiers---in lieu of positional encodings---to learn contextual
embeddings for categorical features with self-attention. Different from TabTransformer, we
view each word location in a sentence as a tabular feature; our
extension thus represents each feature in tabular data via an encoding
akin to the positional encodings. Other tabular extensions of
self-attention include FTTransformer (tokenizing each feature,
applying transformer layers, and using the \texttt{[CLS]} token for
prediction)~\citep{gorishniy2021revisiting}, AutoInt (mapping all
features into the same space and applying self-attention to model
between-feature interactions)~\citep{song2019autoint} and TabNet
(utilizing sequential attention for feature selection in different
learning steps) \citep{arik2020tabnet}. Compared with these existing
approaches, our approach is more robust to covariate shifts across
empirical studies; it also facilitates the integration of heterogeneous
datasets with partially overlapping features.

The third theme relates to \textit{linear word analogy structures in word
embeddings}. Neural word embeddings such as word2vec
\citep{mikolov2013distributed} and GloVe \citep{pennington2014glove}
have been empirically shown to exhibit linear structures as manifested through analogies. Concretely, given an analogy
``\textit{a} is to \textit{b} as \textit{c} is to \textit{d}", we
often find $w_b + w_c - w_a \approx w_d$, where $w_i$ denotes the
embedding of word $i \in \{a,b,c,d\}$. Many works provide theoretical
justifications for this phenomenon. \citet{arora2016a} offered a
latent variable argument, assuming that texts are generated from
random walks of discourse vectors and word vectors are spatially
isotropic. \citet{ethayarajh2018towards} introduced the
\textit{co-occurrence shifted PMI} concept which characterizes when
linear analogy holds in
\gls{SGNS} and GloVe. \citet{allen2019analogies} adopted the
paraphrasing framework of \citet{gittens2017skip} and used
\textit{word transformation} to connect linear analogy in \gls{SGNS}
with paraphrases. In contrast to these existing works, our work moves
beyond  \gls{SGNS} and GloVe; we characterize when linear word
analogies may be present in \gls{CBOW} and attention-based embeddings.


\section{Bidirectional attention as a mixture of continuous word
experts}
\label{sec:bid-attn-CBOW}

In this section, we first review bidirectional attention, a model composed of the self-attention architecture, positional
encodings, and the \gls{MLM} training objective.  En route, we
derive an explicit form of the \gls{MLM} objective for a single-layer
single-head attention-based architecture in \Cref{sec:der-mlm-obj}. We
then formally establish the equivalence between fitting bidirectional
attention and fitting the \gls{CBOW} model with \gls{MoE} weights in
\Cref{sec:mlm-as-cbow}, with extensions to multi-head and multi-layer
attention-based architectures.

\subsection{Bidirectional attention: Self-attention, positional
encodings, and the \gls{MLM} objective}
\label{sec:der-mlm-obj}

We begin with describing the structure of bidirectional
attention---self-attention, positional encodings, and the \gls{MLM}
objective---in the context of language modeling.
(\Cref{sec:app-sum-not} contains a summary of the notations used in
this section.)

\parhead{Building blocks of bidirectional attention.} Consider a
corpus that consists of sentences of length $S$, with a vocabulary
size of $|V|$. The self-attention mechanism takes sentences and
outputs their sentence embeddings, by transforming the token
embeddings and positional encodings of each token in the sentence. We
denote $C \in \mathbb{R}^{(|V| + 1) \times p}$ as the matrix such that
each row $c_i^\top$ corresponds to the token embedding of the $i$-th
token in the vocabulary. The $(|V|+1)$-th token is the \texttt{[MASK]}
token, representing a token in the training corpus that is masked.
We further denote $P \in \mathbb{R}^{S \times p}$ as the positional
encoding matrix.

To learn these token embeddings and positional encodings,
bidirectional direction employs an \gls{MLM} objective: it randomly
masks a random subset of the tokens in the training corpus; then it
aims to predict these masked tokens from the sentence embeddings,
which are produced by transforming the token embeddings and
positional encodings through the attention mechanisms. To
operationalize the \gls{MLM} objective, we use $\overline{X} \in
\{0,1\}^{S \times (|V| + 1)}$ to denote the one-hot encoding matrix of
the $S$ tokens (including the masked tokens) in each sentence. For
notational simplicity, we consider a simple masking strategy: each
sentence produces $S$ prediction tasks in the \gls{MLM} objective,
each of which involves masking exactly one of the $S$ positions in the
sentence and predicting the token in that position. (Results in this
section can be easily generalized to general masking strategies.)

\parhead{Predicting masked tokens with self-attention.} We next
describe how the self-attention mechanism with positional encodings
produces predictions of masked tokens. For ease of exposition, we
focus on a single-head single-layer attention module. It takes in
$\overline{X}$, the one-hot encoding matrix of the $S$ tokens in a
sentence (including the masked tokens); it then outputs a probability
vector $\hat{y}\in
\Delta^{|V|}$ as the prediction for the masked token, indicating the probability of the
masked token being each of the $|V|$ words in the vocabulary.

The self-attention architecture transforms $\overline{X}$ into the
prediction $\hat{y}$ through the following steps:
\begin{enumerate}[leftmargin=*]
    \item \textbf{Token embeddings with positional encodings}. We produce
    a matrix consisting of the token embeddings of all the tokens in the
    masked sentence: $X = \overline{X} C \in \mathbb{R}^{S \times p}$. We 
    then add positional encodings to the matrix: $X' = X + P$.
    \item \textbf{Sentence embeddings with attention weight matrices:}
    Employing the value mapping $W^{V} \in \mathbb{R}^{d \times p}$, query
    mapping $W^{Q} \in
    \mathbb{R}^{d_w \times p}$, and key mapping $W^{K} \in
    \mathbb{R}^{d_w \times p}$, we obtain the sentence embedding
     $X^{\textrm{attn}}\in \mathbb{R}^{S \times d}$ after applying the
     attention weights:
    \[X^{\textrm{attn}} = \textrm{softmax}\left( \frac{X' (W^{Q})^\top W^{K} (X')^\top}{\sqrt{d_w}} \right) X' (W^{V})^\top .\] Here, the softmax is taken row-wise.
    \item \textbf{Intermediate representations with a residual connection.} We then obtain an intermediate representation with the
    coefficient matrix $W^O
    \in \mathbb{R}^{d \times p}$: $Z =
    X^{\textrm{attn}} W^O
    \in
    \mathbb{R}^{S \times p}$. This is followed by a residual connection: $Z' = X' + Z \in \mathbb{R}^{S
    \times p}$. 
    \item \textbf{Final predictions with a linear layer and residual
    connection.} For each position $i \in [S]$ of the sentence, we apply
    a linear layer $\mathrm{LIN}_1(Z_i') = W' Z_i' \in \mathbb{R}^{p}$
    with a weight matrix $W' \in \mathbb{R}^{p \times p}$. We then apply
    another residual connection: $Z'' = Z' + \mathrm{LIN}_1(Z') \in
    \mathbb{R}^{S \times p}$. Finally, we employ another linear layer and the softmax
    operation: \[\hat{y} = \mathrm{softmax}(\mathrm{LIN}_2(Z''_i)),\] where $\mathrm{LIN}_2(Z''_i) = W''
    Z''_i \in \mathbb{R}^{|V|}$ with a weight matrix $W'' \in
    \mathbb{R}^{|V| \times p}$.
\end{enumerate}

Given the self-attention transformation from an input sentence $\overline{X}$ to a probability vector $\hat{y}$ which predicts the masked token, bidirectional
attention learns the token embeddings, positional encodings, and
weight matrices by optimizing the cross entropy loss of $\hat{y}$ in
predicting the masked token. This loss objective is also known as the
\gls{MLM} objective.


\parhead{The loss objective of bidirectional attention.} We next
derive an explicit form for the loss objective of bidirectional
attention. This derivation will pave the road for the statistical
interpretation of bidirectional attention.

In more detail, we consider an input-output pair $(\overline{X},
\overline{y})$ for the masked token prediction task, where
$\overline{X}$ is the one-hot encoding matrix of all tokens in the
sentence, and $\overline{y} \in
\{0,1\}^{|V|}$ is the one-hot encoding of the masked token. We
denote $m \in [S]$ and $b \in [|V|]$ as the masked position and masked
token, respectively. \Cref{prop:mlm-obj} below derives an explicit form
of the \gls{MLM} objective $L_{\gls{MLM}}(m,b)$.

\begin{lemma}[The loss objective of bidirectional attention]
\label{prop:mlm-obj}
Upon reparameterization, the \gls{MLM} objective for predicting token
$b$ in the $m$-th position is given by
\begin{align*}
L_{\gls{MLM}}&(m,b) = - \frac{\sum_{j=1}^S\theta(j,m)\chi(j,m,b) }{\sum_{j=1}^S \theta(j,m)} +\log\left( \sum_{k=1}^{|V|} \exp\left( \frac{\sum_{j=1}^S\theta(j,m)\chi(j,m,k) }{\sum_{j=1}^S \theta(j,m)} \right)\right),
\end{align*}
where
\begin{align*}
\theta(j,m) &\triangleq \exp\left(\frac{ e_j^\top (\overline{X} C+P) W^{KQ} (c_{|V|+1} + P^\top e_m)}{\sqrt{d_w}} \right),\\
\chi(j,m,k) &\triangleq \left( W^{LOV} (\overline{X} C+P)^\top e_j + g + D e_m \right)_k,
\end{align*}
and $g \in \mathbb{R}^{|V|}$, $D \in \mathbb{R}^{|V| \times S}$,
$W^{LOV} \in \mathbb{R}^{|V| \times p}$, $W^{KQ} \in \mathbb{R}^{p
\times p}$, and $e_j \in \{0,1\}^{S}$ denotes a zero vector with 1 on the
$j$-th entry. (The proof is in \Cref{sec:app-proof-mlm-objective}.)
\end{lemma}

\Cref{prop:mlm-obj} performs a reparameterization over the weight
matrices $W^V, W^Q, W^K, W^O$, arriving at an explicit form of the
\gls{MLM} objective with only two weight matrices $W^{KQ}, W^{LOV}$. In particular, $g = W^\ell c_{|V| + 1}$, $D = W^\ell P^\top$, $W^{LOV} = W^\ell (W^O)^\top W^V$, and $W^{KQ} = (W^K)^\top W^Q$ for some weight matrix $W^\ell$. \Cref{prop:mlm-obj} also reveals two key components of the \gls{MLM}
objective: $\theta(j,m)$, the attention weight of token in the $m$-th position on token
in the $j$-th position, and $\chi(j,m,\cdot)$, the similarity between token in the $m$-th position and token
in the $j$-th position. These quantities will play a key role in facilitating the
statistical interpretation of bidirectional attention.

\subsection{Bidirectional attention as a mixture of continuous word
experts}
\label{sec:mlm-as-cbow}

\glsreset{CBOW}
\glsreset{MoE}

Building on the derivations in \Cref{prop:mlm-obj}, we next establish
the equivalence between the loss objective of bidirectional attention
and that of the \gls{CBOW} model with
\gls{MoE} weights. This equivalence will enable us to interpret
bidirectional attention as fitting a statistical model of
\gls{CBOW}+\gls{MoE}.

\parhead{The continuous bag of words model (\gls{CBOW}).} We begin with reviewing
the \gls{CBOW} formulation of word2vec~\citep{mikolov2013distributed}.
\gls{CBOW} aims to predict the center token based on the
surrounding tokens (a.k.a. context tokens). It has two parameter
matrices, representing the center and context embeddings respectively. 

In more detail, we consider an input-output pair $(\overline{X},
\overline{y})$ as in \Cref{sec:der-mlm-obj}, where $m \in [S]$ and $b
\in [|V|]$ represent the masked position and masked token. We note
that, while masking is never employed in \gls{CBOW}, introducing
masking into \gls{CBOW} does not change its objective. The reason is
that the context of a token in \gls{CBOW} does not include the token
itself. Thus, with window size $w$, the loss objective for predicting
the token in the $m$-th position of \gls{CBOW} (a.k.a. the negative
log-likelihood) is
\begin{align*}
L_{\gls{CBOW}}(m,b)=- \frac{\sum_{j=1}^S \omega(j,m) \xi(j, b)}{\sum_{j=1}^S \omega(j,m)} + \log &\left( \sum_{k=1}^{|V|} \exp \left( \frac{\sum_{j=1}^S \omega(j,m) \xi(j, k)}{\sum_{j=1}^S \omega(j,m) } \right)\right),
\end{align*}
where
\begin{align*}
\omega(j,m) &\triangleq \mathds{1}(1 \leq |j-m| \leq w),\\
\xi(j,k) &\triangleq \left( W^{LOV} (\overline{X} C)^\top e_j \right)_k,
\end{align*}

if we denote the center and context embedding matrices by $W^{LOV}$ and~$C$ to
match the notations of bidirectional attention.

\parhead{Weight and similarity matrices in \gls{CBOW} and
bidirectional attention.} The \gls{CBOW} model appears to be related to
bidirectional attention: it admits natural notions of (attention)
\textit{weight} and (token) \textit{similarity} as in bidirectional
attention. Specifically, the \textit{weight} of the token in position
$j \in [S]$ is determined by the distance between $j$ and $m$ and the
number of integers between $m-w$ and $m+w$ (inclusive) that are within
the range $[1,S]$. The \textit{similarity} of token $\alpha \in [|V|]$
in the center and token $\beta \in [|V|]$ in the context is
$(W^{LOV}_\alpha)^\top c_\beta$, regardless of their positions in the
sentence.

To compare \gls{CBOW} and bidirectional attention, we next inspect the
weight matrices in the \gls{MLM} objective of bidirectional attention.
Specifically, the weight of the token in the $j$-th position in
$L_{\gls{MLM}}$ is given by\footnote{The weight and similarity
matrices can take other parametric forms; e.g.,
\citet{sonkar2020attention} uses a different weight function that
depends on the center token $b$ in their \gls{AWE} model.}
\begin{align*}
    \frac{\exp\left(e_j^\top (\overline{X} C+P) W^{KQ} (c_{|V|+1} + P^\top e_m)/\sqrt{d_w}\right)}{\sum_{j=1}^S \exp\left( e_j^\top (\overline{X} C+P) W^{KQ} (c_{|V|+1} + P^\top e_m)/\sqrt{d_w} \right)}.
\end{align*}
Unlike that of \gls{CBOW}, this weight matrix of bidirectional
attention depends on all tokens in the masked sentence and their
corresponding positions. Yet, it does not depend on the center
(masked) token $b$. Further, the term inside the $\exp(\cdot)$ can be
decomposed into four components: 
\begin{enumerate}
    \item $e_j^\top \overline{X} C W^{KQ}
c_{|V|+1}/\sqrt{d_w}$, which depends only on the token in position $j$;
    \item $e_j^\top
\overline{X} C W^{KQ} P^\top e_m/\sqrt{d_w}$, which depends on the token in
position $j$, and position $m$;
    \item  $e_j^\top P W^{KQ} c_{|V|+1}
/\sqrt{d_w}$, which depends only on position $j$;
    \item $e_j^\top P
W^{KQ} P^\top e_m/\sqrt{d_w}$, which depends on both position $j$ and
position $m$. 
\end{enumerate}

The similarity matrix of bidirectional attention also appears to be related
to that of \gls{CBOW}. In bidirectional attention, the similarity of
token $\alpha$ in the center (in position $m$) and token $\beta$ in the
context (in position $j$) is given by $(W_\alpha^{LOV})^\top c_\beta +
(W_\alpha^{LOV})^\top P^\top e_j + g_\alpha + (D_\alpha)^\top e_m$,
which also contains four components. Moreover, the first component coincides with the similarity matrix of \gls{CBOW}.

\parhead{Bidirectional attention as a mixture of continuous word
experts.} Following these observations that bidirectional attention
appears to be closely related to \gls{CBOW}, we conclude this section with
\Cref{prop:mlm-moe-equiv}: it proves that the \gls{MLM} objective of
bidirectional attention in \Cref{prop:mlm-obj} is equivalent to the
\gls{CBOW} objective with \gls{MoE} weights, where the token in each
position serves as an expert.

\glsreset{MoE}

\begin{thm}[Bidirectional attention as a mixture of continuous word experts]
\label{prop:mlm-moe-equiv}
The \gls{MLM} objective of bidirectional attention is equivalent to the cross-entropy loss between the token being masked $\overline{y}$ and the prediction probabilities $\textrm{softmax}(F(\overline{X}))$ from a \gls{MoE} predictor: 
\[F(\overline{X}) = \sum_{j \in [S]} \pi_j(\overline{X}) f_j(\overline{X}),\]
where the $j$th expert $f_j(\overline{X})$ relies on the embedding of
the token in position $j$,
\begin{align*}
    f_j(\overline{X}) = W^{LOV} (\overline{X} C+P)^\top e_j + g + D e_m,
\end{align*}
and its weight (namely the contribution of expert $j$ to the prediction)
is $\pi_j(\overline{X}) =
\left(\mathrm{softmax}(h(\overline{X}))\right)_j$ with
\begin{align*}
    h_j(\overline{X}) = e_j^\top (\overline{X} C+P) W^{KQ} (c_{|V|+1} + P^\top e_m)/\sqrt{d_w}.
\end{align*}
\end{thm}

\Cref{prop:mlm-moe-equiv} is an immediate consequence of
\Cref{prop:mlm-obj}. It formally establishes the equivalence between
bidirectional attention and \gls{CBOW}+\gls{MoE}, enabling a
statistical interpretation of bidirectional attention. In particular,
\Cref{prop:mlm-moe-equiv} reveals the distinct use of \gls{MoE}, a machine learning technique that excels at handling heterogeneous data, in
bidirectional attention. This observation partially explain
the empirical effectiveness of attention-based models in capturing
heterogeneous patterns in complex natural language
data~\citep{devlin2018bert,liu2019roberta}. Moreover, the form of $h_j(\overline{X})$ bears resemblance to a kernel smoother in the sense that it is roughly a weighted inner product between the $j$-th token and all other tokens within the sentence.

\parhead{Extensions to multi-head and multi-layer bidirectional
attention.} We finally extend \Cref{prop:mlm-moe-equiv} to multi-head
and multi-layer bidirectional attention. For bidirectional attention
with multiple attention heads, its \gls{MLM} objective can be shown to
be equivalent to a stacked \gls{MoE} of \gls{CBOW}. For example, for
bidirectional attention with two attention heads, its \gls{MLM}
objective is equivalent to cross entropy loss with the following
stacked \gls{MoE} predictor: 
\[F(\overline{X}) = \sum_{j \in [S]}
\pi^1_j(\overline{X}) f^1_j(\overline{X}) + \sum_{j \in [S]}
\pi^2_j(\overline{X}) f^2_j(\overline{X}),\]
where the $j$th expert of the $i$th head is
\begin{align*}
    f_j^i(\overline{X}) = W^{LOV_i} (\overline{X} C+P)^\top e_j + \frac{g}{2} + \frac{D e_m}{2}
\end{align*}
and the corresponding \gls{MoE} weight is $\pi^i_j(\overline{X}) =
\left(\textrm{softmax}(h^i(\overline{X}))\right)_j$, with
\begin{align*}
    h_j^i(\overline{X}) = e_j^\top (\overline{X} C+P) W^{KQ_i} (c_{|V|+1} + P^\top e_m)/\sqrt{d_w}.
\end{align*} 

Following similar derivations, one can show that bidirectional attention with multiple attention layers is equivalent to a mixture of \glspl{MoE}, i.e., a deep \gls{CBOW} model with weights determined by \glspl{MoE}.

\section{Bidirectional attention for tabular data}

The equivalence between \gls{MLM} with self-attention and \gls{CBOW}
with \gls{MoE} weights (\Cref{prop:mlm-moe-equiv}) suggests an
immediate extension to categorical tabular data, as developed in this section. Across empirical studies, we find
that this tabular extension of attention achieves significant
improvement in \gls{OOD} generalization over existing methods,
including existing algorithms for tabular data (e.g., random forest and
gradient boosting) and existing tabular generalizations of attention
modules (e.g., TabTransformer and FTTransformer).

\subsection{Tabular extension of bidirectional attention}
\label{sec:att-tab}

To extend bidirectional attention to tabular data, we consider a
classification problem with categorical features.  For simplicity, we
assume the response variable $Y_i$ is ordinal with $C$ classes.
Further assume each of the $K$-dimensional features $X_i$ is also
ordinal with $C$ classes. The training data contains pairs of features
and responses $(X_i, Y_i)$. The goal is to predict the response for
some test data point $X$.

Extending bidirectional attention to this tabular setting requires
that we handle tabular features with bidirectional attention. To this
end, we leverage the observation in \Cref{prop:mlm-moe-equiv} that
bidirectional attention can be viewed as \gls{MoE}
with the token in each position (endowed with
positional encodings) serving as an expert. This
\gls{MoE} perspective of bidirectional attention immediately suggests
that we consider each tabular feature as an expert, since
each position in a sentence can be viewed as a tabular feature for
predicting the masked token. To summarize, one can use tabular feature encodings in place of positional encodings for analyzing tabular data with bidirectional attention.

To operationalize this idea,
we first introduce ``word" embeddings $w_1, \cdots, w_C \in
\mathbb{R}^d$ for each class and $w_0$ for the \texttt{[MASK]} token.
We then introduce ``position" encodings $p_1, \cdots, p_{K+1} \in
\mathbb{R}^d$, one for each feature. Finally, we consider the
concatenation of features and covariates $(X_i, Y_i)$ of each data
point as a sentence in bidirectional attention. In order to learn the word embeddings and positional encodings using the \gls{MLM} objective, we do the following:

\begin{enumerate}
    \item Replace each position with the \texttt{[MASK]} token, one at a time.
    \item For each masking instance, compute the sum of the corresponding word embeddings and positional encodings, and employ the usual cross-entropy loss to predict the masked token.
    \item At test time, given a test data point $X_{\textrm{test}}$, use the trained model to predict the most probable class for the input $(X_{\textrm{test}}, \texttt{[MASK]})$.
\end{enumerate}

We note that this use of \gls{MLM} objective for tabular data
implicitly models the joint distribution $p(X,Y)$, as opposed to the
conditional distribution $p(Y|X)$ that standard supervised algorithms
commonly model. As a consequence, tabular extensions of bidirectional
attention can potentially achieve better \gls{OOD} generalization, as we demonstrate
empirically next.


\begin{table}[h!]
\centering
\label{fig:sim-res-acc}
\begin{subtable}[t]{\linewidth}
\centering
\begin{tabular}{cccccc}
\hline
Param. $\backslash$ \textbf{Acc.} & LR & RF & GB & MLP & ATN \\ \hline
$(1, 0, 0.1)$ & 0.388 & 0.409 & \textbf{0.413} & 0.323 & 0.404 \\ 
$(1, 0, 0.9)$ & 0.313 & 0.298 & 0.350 & 0.237 & \textbf{0.389} \\ \hline
$(1, 0.5, 0.1)$ & 0.345 & 0.361 & \textbf{0.366} & 0.292 & 0.359 \\ 
$(1, 0.5, 0.9)$ & 0.270 & 0.253 & 0.299  & 0.202 & \textbf{0.306} \\ \hline
$(1, 1.5, 0.1)$ & 0.250 & 0.243 & \textbf{0.253} & 0.204 & 0.252 \\ 
$(1, 1.5, 0.9)$ & 0.169 & 0.158 & \textbf{0.172} & 0.142 & 0.170 \\ \hline
$(5, 0, 0.1)$ & 0.250 & 0.207 & 0.244 & 0.306 & \textbf{0.419} \\ 
$(5, 0, 0.9)$ & 0.162 & 0.150 & 0.156 & 0.169 & \textbf{0.392} \\ \hline
$(5, 0.5, 0.1)$ & 0.227 & 0.173 & 0.214 & 0.252 & \textbf{0.318} \\ 
$(5, 0.5, 0.9)$ & 0.154 & 0.133 & 0.153 & 0.151 & \textbf{0.269} \\ \hline
$(5, 1.5, 0.1)$ & 0.167 & 0.099 & 0.157 & 0.165 & \textbf{0.171} \\ 
$(5, 1.5, 0.9)$ & 0.125 & 0.108 & 0.114 & 0.118 & \textbf{0.133} \\ \hline
\end{tabular}
\caption{Accuracy}
\end{subtable}
\vspace{15pt}
\begin{subtable}[t]{\linewidth}
\centering
\begin{tabular}{cccccc}
\hline
Param. $\backslash$ \textbf{MSE} & LR & RF & GB & MLP & ATN \\ \hline
$(1, 0, 0.1)$ & 3.015 & \textbf{2.694} & 2.730 & 4.059 & 2.941 \\ 
$(1, 0, 0.9)$ & 5.163 & 9.331 & 4.855 & 7.911 & \textbf{3.078} \\ \hline
$(1, 0.5, 0.1)$ & 3.416 & 3.201 & \textbf{3.123} & 4.704 & 3.281 \\ 
$(1, 0.5, 0.9)$ & 5.955 & 10.106 & 6.070 & 8.123 & \textbf{4.465} \\ \hline
$(1, 1.5, 0.1)$ & 5.725 & 5.685 & \textbf{5.415} & 7.199 & 5.594 \\ 
$(1, 1.5, 0.9)$ & 8.942 & 12.340 & 9.837 & 9.874 & \textbf{7.339} \\ \hline
$(5, 0, 0.1)$ & 5.333 & 8.498 & 5.967 & 2.814 & \textbf{1.521} \\ 
$(5, 0, 0.9)$ & 5.674 & 10.101 & 8.858 & 7.842 & \textbf{1.633} \\ \hline
$(5, 0.5, 0.1)$ & 6.021 & 10.236 & 6.844 & 4.056 & \textbf{2.605} \\ 
$(5, 0.5, 0.9)$ & 6.118 & 10.427 & 8.283 & 7.884 & \textbf{2.355} \\ \hline
$(5, 1.5, 0.1)$ & 9.159 & 16.154 & 9.538 & \textbf{8.313} & 8.316 \\ 
$(5, 1.5, 0.9)$ & 8.410 & 10.409 & 10.110 & 9.966 & \textbf{6.501} \\ \hline
\end{tabular}
\caption{MSE}
\end{subtable}
\caption{The proposed tabular extension of bidirectional attention
(ATN) achieves better or competitive accuracy and MSE than competing
methods, across all parameter settings. The parameter tuples indicate
different choices of $(n_c, \textrm{noise}, \textrm{corr})$.}
\end{table}

Finally, this tabular extension of bidirectional attention can be
applied beyond supervised classification. It readily extends to
unsupervised settings (if we ignore the $Y_i$'s) and semi-supervised
settings (if we consider both the labeled and unlabeled data and set
the $Y_i$'s for the unlabeled data to be \texttt{[MASK]}). This
approach is also applicable to multiple data sets with only
partially overlapping features: the learned feature encodings will
allow us to bring all features into the same embedding space. These
learned encodings can also reveal the relationships between different
tabular features across different data sets.

\subsection{Empirical studies of tabular bidirectional attention}
\label{sec:att-tab-emp}

In this section, we empirically study the tabular extension of
bidirectional attention using simulated and real data sets. Across
empirical studies, we find that this approach outperforms in
\gls{OOD} generalization for tabular data, as compared with both
existing tabular data algorithms and existing tabular extensions of attention modules.

\subsubsection{Simulated data}
\label{sec:sim-data}
We begin with evaluating tabular bidirectional attention on simulated data. We
focus on the common \gls{OOD} generalization setting of covariate
shift; it refers to prediction tasks where $p(X_{\textrm{train}}) \neq
p(X_{\textrm{test}})$ and $p(Y_{\textrm{train}} | X_{\textrm{train}})
= p(Y_{\textrm{test}} | X_{\textrm{test}})$.

\parhead{Data generation.} We describe the key components of the data
generation process; we refer the readers to
\Cref{sec:app-tab-data-exp} for full details. We set the number of
features $K$ to be 5, the number of classes $C$ to be 10, and both the
training and test set size to be 2,000. Twenty data sets are
generated for each combination of hyperparameters.

\parhead{Competing methods and evaluation metrics.} We fit the proposed tabular
extension of bidirectional attention model to each training set,
together with a few competing methods, namely logistic regression
(LR), random forests (RF), gradient boosting (GB) and multilayer
perceptron (MLP). See \Cref{app:hyp-tun} for implementation details.

\parhead{Results.} \Cref{fig:sim-res-acc} summarizes the test accuracy
and mean squared error of all methods. We find that the proposed
tabular extension of bidirectional attention outperforms or
competitively compares to all competing methods. Moreover, its
performance gain is more apparent when $\textrm{corr} = 0.9$ (very
correlated training features) as compared to when $\textrm{corr} =
0.1$; the former corresponds to a more challenging case of covariate
shift.

\subsubsection{UCI's auto-mpg data}

We next study the tabular extension of bidirectional attention on a
real data set, namely the \texttt{auto-mpg} data from the UCI data set.
This data set contains the following information from 398 different
car models: \textit{mpg}, \textit{cylinders}, \textit{displacement},
\textit{horsepower}, \textit{weight}, \textit{acceleration},
\textit{model year}, \textit{origin}, and \textit{car name}.

\parhead{Data processing.} To simulate covariate shift, we follow the
approach of \citet{storkey2006mixture}: we assigns cars from origin 1
to the training set, and origins 2 and 3 to the test set. In addition,
we only consider cars with 4, 6 or 8 cylinders and remove data points
with missing values. Lastly, similar to the synthetic data
experiments, we convert each column into three quantile-based
categories. The final data set has 385 data points, where 245 belong
to the training set and 140 belong to the test set.


\begin{table}[ht]
\centering
\label{fig:auto-mpg-res}
\begin{tabular}{ccccccccccc}
\hline
 & LR & RF & GB & MLP  & CE & FT & TT & AI & TN & ATN (ours)\\ \hline
Accuracy & 0.657 & 0.721 & 0.657 & 0.700  & 0.764 & 0.707 & 0.707 & 0.364 & 0.600 & \textbf{0.793}\\ \hline
MSE & 0.343 & 0.279 & 0.343 & 0.300  & 0.236 & 0.293 & 0.293 & 0.636 & 0.486 & \textbf{0.207}\\ \hline
\end{tabular}
\caption{The proposed tabular extension of attention (ATN) achieves
superior performance as compared to all baselines. (Lower MSE and
higher accuracy is better.)}
\end{table}

\parhead{Competing methods and evaluation metrics.} We use the same
competing methods and evaluation metrics as in \Cref{sec:sim-data}.
Additionally, we compare with other existing tabular extensions of
attention modules, including CategoryEmbedding (CE)
\citep{joseph2021pytorch}, FTTransformer (FT)
\citep{gorishniy2021revisiting}, TabTransformer (TT)
\citep{huang2020tab}, AutoInt (AI) \citep{song2019autoint}, and TabNet
(TN) \citep{arik2020tabnet}.\footnote{We use
\texttt{pytorch\_tabular}'s \citep{joseph2021pytorch} implementation
with the default parameters. The batch and epoch sizes are set to be
128 and 200, respectively.}

\parhead{Results.} \Cref{fig:auto-mpg-res} summarizes the test
accuracy and mean squared error of all methods. We find that the
proposed tabular extension of bidirectional attention outperforms all
competing methods. This performance gain is likely due to its focus on
modeling the joint distribution of the covariates and response
variable; it is in contrast to the practice of modeling only the
conditional distribution of the response variable given the covariates
in supervised learning.


\section{Linear word analogies in attention-based embeddings}
\label{sec:lin-rel-anal}

In this section, we explore the presence of linear word analogies in
the embeddings of bidirectional attention and its non-attention
predecessors. En route, we leverage the close connection between
\gls{CBOW} and bidirectional attention in \Cref{prop:mlm-moe-equiv} to
facilitate the theoretical analysis. This exploration is motivated by
a curious empirical observation: While bidirectional attention (e.g.,
BERT) often significantly outperforms its non-attention predecessors
in natural language processing benchmarks, it does not seem to
outperform its predecessors in word analogy tasks. In particular, it
can sometimes perform worse as compared to classical word
embedding algorithms like word2vec~\citep{mikolov2013distributed} and
GloVe~\citep{pennington2014glove}.

Thanks to these empirical observations, we characterize under which
conditions bidirectional attention and \gls{CBOW} can exhibit linear
word analogies in their embeddings. We find that bidirectional
attention requires much stronger conditions to exhibit linear word
analogies than its non-attention predecessors. These results partially
explain the limited empirical gain in using bidirectional attention
for word analogy tasks.

\subsection{A curious empirical study: Do attention-based token embeddings
exhibit linear word analogies?}
\label{sec:attn-lin-emp}

We begin with a curious empirical study about the presence of linear
word analogies in attention-based and non-attention-based token
embeddings. Linear structure in neural word embeddings such as
word2vec \citep{mikolov2013distributed} and GloVe
\citep{pennington2014glove} is a well-known empirical phenomenon.
However, most studies focused on embeddings trained via
\gls{SGNS}~\citep{ethayarajh2018towards,allen2019analogies}. This
phenomenon is less studied in other language modeling
approaches, e.g., \gls{CBOW} and bidirectional attention, with few
exceptions~\citep{ushio2021bert}.

To this end, we first perform an empirical study about whether linear
relationships are observed in embeddings from word2vec trained with
the \gls{CBOW} objective and BERT \citep{devlin2018bert}, a large language
model based on bidirectional attention. Following existing
studies, we use the analogy identification task as a proxy for
identifying the presence of linear relationships, utilizing the analogy
data set first introduced in \citet{pennington2014glove}. We refer the
readers to \Cref{sec:analogy-details-expm} for data set and
implementation details.

\parhead{Evaluation metrics.} For each model, we are interested in (1) the overall and per-category accuracies, where accuracy is defined as the proportion of correct answers; and (2) the overall and per-category average cosine similarity between $x_b + x_c - x_a$ and the correct answer. We note that (2) is a better metric than (1) due to the difference in vocabulary sizes across models.

\parhead{Results.} The accuracy and average cosine similarity for each
model is displayed in \Cref{table:acc}. We observe that all three
models generally result in word embeddings that exhibit certain degrees of linear
word analogies. However, the bidirectional attention model BERT can
often perform worse than its non-attention predecessor GloVe in this
task, despite it being a much more powerful language model as evidenced by its performance in common natural language benchmarks. 

What factors have limited BERT's (and bidirectional attention's)
ability to exhibit linear word analogies? What about \gls{CBOW} and
GloVe? Below we study these questions theoretically, leveraging the
close connection between \gls{CBOW} and bidirectional attention in
\Cref{prop:mlm-moe-equiv}. In particular, we characterize the
conditions under which \gls{CBOW} and bidirectional attention may
exhibit linear word analogies. We find that the
conditions required by bidirectional attention are much stronger, which
partially explains the empirical observations above.


\begin{table*}[t]
\centering

\label{table:acc}
\begin{tabular}{lllll}
\hline
\textbf{Accuracy}  & BERT  & GloVe & CBOW \\ \hline
Semantic  & 0.641 & 0.759 & 0.234 \\ \hline
Syntactic & 0.754 & 0.692 & 0.667 \\ \hline
Overall   & 0.727 & 0.708 & 0.563 \\ \hline
\end{tabular}
\qquad
\begin{tabular}{lllll}
\hline
\textbf{Cosine similarity}  & BERT  & GloVe  & CBOW \\ \hline
Semantic  & 0.500 & 0.600 & 0.504    \\ \hline
Syntactic & 0.610 & 0.610 & 0.582     \\ \hline
Overall   & 0.584 & 0.607 & 0.564     \\ \hline
\end{tabular}
\caption{Classical word embedding methods can achieve similar or
higher performance than attention-based model in word analogy tasks:
GloVe achieve higher or the same average cosine similarity than BERT
on both syntactic and semantic analogies; GloVe also outperforms BERT
in accuracy for semantic analogies. (Higher is better.)}
\end{table*}

\subsection{Linear word analogies in CBOW and bidirectional attention
embeddings}
\label{sec:lin-rel-cbow}

We begin with theoretically characterizing under which conditions
\gls{CBOW} embeddings can exhibit linear word analogies. Starting with
\citeauthor{allen2019analogies}'s [\citeyear{allen2019analogies}]
argument for \gls{SGNS}, we extend the argument to both \gls{CBOW} and
attention-based token embeddings, thanks to the equivalence we
established in \Cref{prop:mlm-moe-equiv}.

To perform this theoretical analysis, we follow existing analyses about \gls{SGNS}: \citet{levy2014neural} showed that for a sufficiently large embedding dimension, embeddings from \gls{SGNS} satisfy
    $w_i^\top c_j = \log \left( \frac{p(w_i, c_j)}{p(w_i) p(c_j)}\right) - \log k = \textrm{PMI}(w_i, c_j) - \log k,$
where $k$ is the number of negative samples for each positive sample;
$W^{LOV}, C \in \mathbb{R}^{|V| \times p}$ are the center and context
embedding matrix, respectively. For each $i \in [|V|]$, $w_i^\top$
($c_i^\top$) is the $i$-th row of $W^{LOV}$ ($C$), which represents
the center (context) embedding of word $i$.

Using this result, \citet{allen2019analogies}
considered embeddings which factorize the unshifted PMI matrix, namely
$w_i^\top c_j = \textrm{PMI}(w_i, c_j)$, compactly written as $W^\top
C = \textrm{PMI}$. Through the ideas of \textit{paraphrases} and
\textit{word transformations}, they explained why linear relationships
exist for analogies on \gls{SGNS} word embeddings.

Here, we perform similar analyses for \gls{CBOW} and bidirectional
attention; the goal is to characterize the conditions under which
\gls{CBOW} and bidirectional attention can exhibit linear word
analogies respectively. Below we sketch the main results we obtain,
deferring full details to \Cref{sec:detailed-analogy}.

\parhead{Linear word analogies in \gls{CBOW} embeddings.} We first
characterize the inner product of center and context embeddings of
\gls{CBOW}.
\begin{prop}
\label{prop:cbow-sim}
Embeddings from fitting \gls{CBOW} without negative sampling must satisfy
    $w_i^\top c_j \approx \log \left( \frac{p(w_i, c_j)}{p(c_j)} \right) + \log |V|.$
\end{prop}
This result suggests that \gls{CBOW} approximately factorizes $M$, a
$|V| \times |V|$ matrix such that $M_{i,j} = \log \left(\frac{p(w_i,
c_j)}{p(c_j)} \right) + \log |V|.$ Following this result, we next
argue that the \gls{CBOW} embeddings approximately form a linear
relationship, up to some error terms.
\begin{prop}
\label{prop:anal-w2v}
    Given any $w_a, w_{a^*}, w_b, w_{b^*} \in \mathcal{E}$, we have
    \begin{align*}
            w_{b^*}
            &= w_{a^*} - w_a + w_b + C^{\dagger}(\rho^{\mathcal{W}, \mathcal{W}_*} + \Delta^{\mathcal{W}, \mathcal{W}_*}+ \delta^{\mathcal{W}, \mathcal{W}_*}) \\
            &= w_{a^*} - w_a + w_b + C^{\dagger}(\xi^{\mathcal{W}, \mathcal{W}_*} + \Delta^{\mathcal{W}, \mathcal{W}_*}),
    \end{align*}
    where $\mathcal{E}$ is the set of all words in the vocabulary,
    $\mathcal{W} = \{ w_b, w_{a^*}\}$, $\Delta^{\mathcal{W},
    \mathcal{W}_*}=\sigma^{\mathcal{W}} - \sigma^{\mathcal{W}_*} $ and
    $\mathcal{W}_* = \{ w_{b^*}, w_a \}$. The quantities
    $\rho^{\mathcal{W},
    \mathcal{W}_*}, \Delta^{\mathcal{W}, \mathcal{W}_*},
    \delta^{\mathcal{W}, \mathcal{W}_*}, \xi^{\mathcal{W},
    \mathcal{W}_*}$ are all statistics that characterize the
    relationships between the two word sets $\mathcal{W},
    \mathcal{W}_*$. We refer the reader to \Cref{sec:detailed-analogy}
    for their precise definitions and complete details of the results.
\end{prop}

\Cref{prop:anal-w2v} reveals that we have linear word analogies
$w_{b^*} \approx w_{a^*} - w_a + w_b$ when $\mathcal{W}$ paraphrases
$\mathcal{W}_*$ in the sense of \citet{allen2019analogies} (i.e.
$\rho^{\mathcal{W}, \mathcal{W}_*}\approx 0$), and
$\sigma^{\mathcal{W}}$, $\sigma^{\mathcal{W}_*}$ and
$\delta^{\mathcal{W}, \mathcal{W}_*}$ are small. The latter conditions
hold true only when all $w_i \in \mathcal{W}$ ($w_i \in
\mathcal{W}_*$) are approximately conditionally independent given
$c_j$, and $p(\mathcal{W}) \approx p(\mathcal{W}_*)$. If we consider
alternative definitions of paraphrase---which we detail in
\Cref{sec:detailed-analogy}, then the linear analogy error may only
depend on the approximate conditional independence of $w_i$'s
given~$c_j$.

Finally, we characterize the conditions under which, if token embeddings
of \gls{CBOW} exhibit linear word analogies, then its contextual
embedding will also exhibit this structure.
\begin{prop}
\label{prop:anal-ctx}
    Let $\mathcal{W} = \{r,s\}$ and $\mathcal{W}_* = \{t,u\}$. Assume
    $p(\mathcal{W}) \approx p(\mathcal{W_*})$ and $w_i \in
    \mathcal{W}$ ($w_i \in \mathcal{W_*}$) are approximately
    marginally independent. Further, assume that $W$ has full row
    rank. If $w_r + w_s \approx w_t + w_u$, then $c_r + c_s \approx
    c_t + c_u$.
\end{prop}

\parhead{Linear word analogies for bidirectional attention.} We next
extend these \gls{CBOW} arguments to bidirectional attention,
leveraging the close connection established in
\Cref{prop:mlm-moe-equiv}. We will show that the same linear word
analogies may emerge in bidirectional attention, but under much
stronger assumptions.

\begin{prop}
\label{prop:sim-attn}
    Token embeddings from bidirectional attention must satisfy
    \begin{equation*}
    \begin{split}
        w_i^\top c_j 
         &\approx \frac{|V| \sum_{(i,j)} \gamma_{j}^i - \left(\sum_{(1,j)} \gamma_{j}^1 + \cdots + \sum_{(|V|,j)} \gamma_{j}^{|V|} \right)}{S \left(\sum_{(1,j)} (\gamma_{j}^1)^2 + \cdots + \sum_{(|V|,j)} (\gamma_{j}^{|V|})^2 \right)},
    \end{split}
    \end{equation*}
where for a center-context pair $(d,j)$ in the masked sentence $(a_1, \cdots, a_S)$, we define $\gamma_j^d = \tau_j / \sum_{s=1}^S \tau_{a_s}$, and $\tau_j = \exp\left( c_j ^\top W^{KQ} c_{|V|+1}/\sqrt{d_w} \right)$.
\end{prop}
\Cref{prop:sim-attn} shows that bidirectional attention approximately
factorizes a $|V| \times |V|$ matrix whose $(i,j)$-th entry is given
by the equation above. Unlike in \gls{CBOW}, the token embedding for
each word $i$ is $c_i$ (the \textit{context} embedding), and not $w_i$
(the \textit{center} embedding). In the case where $\tau_j$ is
approximately the same for every $j \in [|V|+1]$, the problem
approximately reduces to a vanilla \gls{CBOW}: we always have
$\gamma_j^d \approx 1/S$, whence \Cref{prop:sim-attn} yields $w_i^\top
c_j \approx \frac{p(w_i, c_j)}{p(c_j)} \cdot |V| - 1 \approx \log
\left( \frac{p(w_i, c_j)}{p(c_j)} \right) + \log |V|$.

Following a similar argument as \Cref{prop:anal-w2v}, we argue that the bidirectional attention embedding can also exhibit linear word analogies, up to some error.

\begin{prop}
\label{prop:anal-w2v-attn}
    Given any $w_a, w_{a^*}, w_b, w_{b^*} \in \mathcal{E}$, we have
    \begin{align*}
            w_{b^*} 
            &= w_{a^*} - w_a + w_b + \tilde{C}^{\dagger}(\bar{\rho}^{\mathcal{W}, \mathcal{W}_*} + \overline{\Delta}^{\mathcal{W},
    \mathcal{W}_*} + \bar{\delta}^{\mathcal{W}, \mathcal{W}_*}) \\
            &= w_{a^*} - w_a + w_b + \tilde{C}^{\dagger}(\bar{\xi}^{\mathcal{W}, \mathcal{W}_*} + \overline{\Delta}^{\mathcal{W},
    \mathcal{W}_*}),
    \end{align*}
    where $\overline{\Delta}^{\mathcal{W},
    \mathcal{W}_*}=\overline{\sigma}^{\mathcal{W}}- \overline{\sigma}^{\mathcal{W}_*}$, $\mathcal{W} = \{ w_b, w_{a^*}\}$, and $\mathcal{W}_* = \{ w_{b^*}, w_a \}$. The quantities $\bar{\rho}^{\mathcal{W}, \mathcal{W}_*}$, $\overline{\Delta}^{\mathcal{W}
    \mathcal{W}_*}$, $\bar{\delta}^{\mathcal{W}, \mathcal{W}_*}$ characterize the relationships between $\mathcal{W},
    \mathcal{W}_*$ based on $\bar{p}(w_i, c_j) \triangleq
    \sum_{(i,j)}\gamma_j^i/E$; see details in \Cref{sec:detailed-analogy}.
\end{prop}

Under additional conditions, similar linear word analogy relationships
may also emerge for the contextual embeddings of bidirectional
attention.

\begin{prop}
\label{prop:anal-ctx-attn}
    Let $\mathcal{W} = \{r,s\}$ and $\mathcal{W}_* = \{t,u\}$. Assume
    $\bar{p}(\mathcal{W}) \approx \bar{p}(\mathcal{W_*})$ and $w_i \in
    \mathcal{W}$ ($w_i \in \mathcal{W_*}$) are approximately
    marginally independent. Further assume that $W$ has full row rank
    and $\bar{p}(w_i, c_j) \approx \bar{p}(w_j,c_i)$. If $w_r + w_s
    \approx w_t + w_u$, then $\tilde{c}_r + \tilde{c}_s \approx
    \tilde{c}_t + \tilde{c}_u$.
\end{prop}

While we leave the full details of these results to \Cref{sec:detailed-analogy}, \Cref{prop:anal-w2v-attn,prop:anal-ctx-attn} suggest that bidirectional attention requires much stronger conditions to exhibit linear relationships than \gls{CBOW}. Specifically, it requires the quantity $\bar{p}(w_i, c_j) = \sum_{(i,j)}\gamma_j^i/E$ to be approximately symmetric. Even when this condition holds, linear word analogy would only hold for some transformed embeddings $\tilde{c}_i$'s, as opposed to the token embeddings $c_i$'s. Only under an additional assumption that
   $\zeta_j := \frac{\sum_{(1,j)} (\gamma_{j}^1)^2 + \cdots + \sum_{(|V|,j)} (\gamma_{j}^{|V|})^2 }{\sum_{(1,j)} \gamma_{j}^1 + \cdots + \sum_{(|V|,j)} \gamma_{j}^{|V|} }$
is approximately the same for each $j$ (e.g., when $\tau_j$ is
approximately the same for every $j$) will we approximately have
linear word analogies for the token embeddings $c_i$'s.

Finally, we note that all these results can be easily extended to
incorporate positional encodings by considering each (word, position)
pair as a unit. In these cases, analogies will be drawn between (word,
position) pairs.

\glsresetall

\section{Discussion}

In this paper, we prove that a single-head single-layer bidirectional
attention is equivalent to a \gls{CBOW} model with \gls{MoE} weights,
upon reparameterization. This statistical perspective reveals the
distinct use of \gls{MoE} in bidirectional attention, supporting the
empirical observation that bidirectional attention excels in
capturing heterogeneous patterns. This connection further suggests
immediate extensions of bidirectional attention to tabular data, leading to improved
\gls{OOD} generalizations when compared to existing approaches. It also
allows us to characterize the conditions under which embeddings from
bidirectional attention and \gls{CBOW} exhibit linear word analogies.
These analyses show that bidirectional attention requires much
stronger assumptions than its non-attention predecessors to exhibit
linear word analogies.

One limitation of this work is that the linear word analogy argument
in \Cref{sec:lin-rel-anal} ignores residual connections. In addition,
we only consider bidirectional attention architectures that use linear
layers, as opposed to feed-forward layers used
in~\citet{devlin2018bert}. Beyond addressing these limitations,
exploring the statistical properties of bidirectional attention is an
interesting avenue for future work. It will also be useful to provide
theoretical justifications for the observed robustness of
bidirectional attention to covariate shifts, and to understand the
fundamental differences between static and contextual word embeddings
in their abilities to form linear analogies.

\vspace{20pt}

\parhead{Acknowledgments. } This work
was supported in part by the Office of Naval Research under grant
number N00014-23-1-2590 and the National Science Foundation under
grant numbers 2231174 and 2310831. We thank Sasha Rush for suggesting
the name ``bidirectional attention.''

\clearpage


\bibliography{attention}

\clearpage


\appendix

\onecolumn

\begin{center}
\Large{Supplementary Material: Bidirectional Attention as a Mixture of Continuous Word Experts}
\end{center}

\section{Summary of notations}
\label{sec:app-sum-not}
Below is a summary of commonly-used notations in \Cref{sec:lin-rel-anal}. 

\begin{table}[h]
\centering
\captionsetup{format=empty,aboveskip=0pt,belowskip=0pt}
\begin{tabular}{c|l}
Notation & Explanation \\[0.5mm] \hline
$|V|$ & Vocabulary size \\[0.5mm] 
$S$ & Sentence length \\[0.5mm] 
$p$ & Embedding dimension \\[0.5mm] 
$W^{LOV}$ & Center embedding matrix \\[0.5mm]
$C$ & Token (context) embedding matrix \\[0.5mm] 
$w_i^\top$ & $i$-th row of $W^{LOV}$ \\[0.5mm]
$c_i^\top$ & $i$-th row of $C$ \\[0.5mm]
$P$ & Position encoding matrix \\[0.5mm] 
$\overline{X}$ & One-hot encoding matrix of the masked sentence \\[0.5mm]
$\overline{y}$ & One-hot encoding of the target word \\[0.5mm]
$m$ & Position of the masked word \\[0.5mm] 
$b$ & The masked word \\[0.5mm] 
$e_j$ & A zero vector of length $S$ with 1 on the $j$-th entry \\[0.5mm] 
$f_j(\cdot)$ & The output generated by expert $j$ \\[0.5mm] 
$\pi_j(\cdot)$ & The contribution of expert $j$ \\[0.5mm] 
$a_s$ & The word on the $s$-th position of the masked sentence \\[0.5mm] 
\end{tabular}
\end{table}

\section{A sketch of the attention-based architecture}
\label{sec:app-simple-attn}
\begin{enumerate}
    \item Let $X = \overline{X} C \in \mathbb{R}^{S \times p}$ be a matrix consisting of the token embeddings of each word in the masked sentence, and $X' = X + P \in \mathbb{R}^{S \times p}$. 
    \item Introduce attention weight matrices $W^{V} \in \mathbb{R}^{d \times p}$, $W^{Q} \in \mathbb{R}^{d_w \times p}$ and $W^{K} \in \mathbb{R}^{d_w \times p}$. Let $X^{\textrm{attn}} = \textrm{softmax}\left( \frac{X' (W^{Q})^\top W^{K} (X')^\top}{\sqrt{d_w}} \right) X' (W^{V})^\top \in \mathbb{R}^{S \times d}$, where the softmax is taken row-wise.
    \item Let $W^O \in \mathbb{R}^{d \times p}$, and write $Z = X^{\textrm{attn}} W^O \in \mathbb{R}^{S \times p}$.
    \item Introduce a residual connection, and write $Z' := X' + Z \in \mathbb{R}^{S \times p}$.
    \item For each position $i \in [S]$, apply a linear layer $LIN_1(Z_i') = W' Z_i' \in \mathbb{R}^{p}$, where $W' \in \mathbb{R}^{p \times p}$.
    \item Introduce another residual connection, and write $Z'' = Z' + LIN_1(Z') \in \mathbb{R}^{S \times p}$.
    \item For each position $i \in [S]$, apply a linear layer $LIN_2(Z''_i) := W'' Z''_i \in \mathbb{R}^{|V|}$, where $W'' \in \mathbb{R}^{|V| \times p}$.
    \item Perform the softmax operation and calculate the cross-entropy loss corresponding to predicting the masked word in the sentence.
\end{enumerate}

\section{Proof of Lemma \ref{prop:mlm-obj}}

\label{sec:app-proof-mlm-objective}

\textit{Proof.} Recall that $m \in [S]$ and $b \in [|V|]$ represent the masked position and masked word, respectively. It is easy to see that $X^\top e_m = c_{|V| + 1}$, where $e_m \in \{0,1\}^{S}$ is a zero vector with 1 on the $m$-th entry. Note that steps 1 to 4 of \Cref{sec:app-simple-attn} give us
\begin{equation*}
    Z' = X + P + \textrm{softmax} \left( \frac{(X+P)(W^Q)^\top W^K (X+P)^\top}{\sqrt{d_w}} \right) (X+P) (W^V)^\top W^O \in \mathbb{R}^{S \times p}.
\end{equation*}
This is followed by steps 5 and 6, which yield $Z'' = Z' + LIN_1(Z')$ where the $i$-th row of $Z''$ is given by $(Z_i'')^\top$, where $Z''_i = Z_i' + W' Z_i'$ for some $W' \in \mathbb{R}^{p \times p}$.  Lastly, steps 7 and 8 result in $\alpha_m = \textrm{softmax} (W'' Z_m'')$ for some $W'' \in \mathbb{R}^{|V| \times p}$, from which the loss is simply $-\log(e_b^\top \alpha_m)$, where $e_b \in \{0,1\}^{|V|}$ is a zero vector with 1 on the $b$-th entry. See that 
\begin{align*}
        &W'' Z_m'' \\
        &= (W'' + W'' W')(Z')^\top e_m \\
        &= W^\ell \left( (X + P)^\top + (W^O)^\top W^V  (X + P)^\top \textrm{softmax} \left( \frac{(X+P)(W^K)^\top W^Q (X+P)^\top}{\sqrt{d_w}} \right)  \right) e_m,
\end{align*}
where $W^\ell = W'' + W'' W' \in \mathbb{R}^{|V| \times p}$ and the softmax is taken column-wise. Writing $W^\ell c_{|V| + 1} = g \in \mathbb{R}^{|V|}$, $W^\ell P^\top = D \in \mathbb{R}^{|V| \times S}$, $W^\ell (W^O)^\top W^V = W^{LOV} \in \mathbb{R}^{|V| \times p}$ and $(W^K)^\top W^Q = W^{KQ} \in \mathbb{R}^{p \times p}$, we obtain
\begin{align*}
        W'' Z_m'' &=  g + D e_m + \sum_{j=1}^S \frac{\exp\left( \frac{e_j^\top (X+P) W^{KQ} (c_{|V|+1} + P^\top e_m)}{\sqrt{d_w}} \right)}{\sum_{j=1}^S \exp\left( \frac{e_j^\top (X+P) W^{KQ} (c_{|V|+1} + P^\top e_m)}{\sqrt{d_w}} \right)} \left( W^{LOV} (X+P)^\top e_j\right) \\
        &= \sum_{j=1}^S \frac{\exp\left( \frac{e_j^\top (\overline{X} C+P) W^{KQ} (c_{|V|+1} + P^\top e_m)}{\sqrt{d_w}} \right)}{\sum_{j=1}^S \exp\left( \frac{e_j^\top (\overline{X} C+P) W^{KQ} (c_{|V|+1} + P^\top e_m)}{\sqrt{d_w}} \right)} \left( W^{LOV} (\overline{X} C+P)^\top e_j + g + D e_m \right),
\end{align*}
and the objective for this particular instance is
$$- \frac{\sum_{j=1}^S\theta(j,m)\chi(j,m,b) }{\sum_{j=1}^S \theta(j,m)} +\log\left( \sum_{k=1}^{|V|} \exp\left( \frac{\sum_{j=1}^S\theta(j,m)\chi(j,m,k) }{\sum_{j=1}^S \theta(j,m)} \right)\right)$$
where
\begin{align*}
\theta(j,m) &\triangleq \exp\left(\frac{ e_j^\top (\overline{X} C+P) W^{KQ} (c_{|V|+1} + P^\top e_m)}{\sqrt{d_w}} \right),\\
\chi(j,m,k) &\triangleq \left( W^{LOV} (\overline{X} C+P)^\top e_j + g + D e_m \right)_k,
\end{align*}
completing the proof.

\section{Tabular data generation process}
\label{sec:app-tab-data-exp}

We set the number of
features $K$ to be 5, the number of classes $C$ to be 10, and the
training and test set size to be 2,000 each. Twenty data sets are
generated for each combination of hyperparameters: (1) $n_c \in
\{1,5\}$, the number of features which generate $Y$; (2)
$\textrm{noise} \in \{0, 0.5, 1.5\}$, where a larger value indicates a
larger noise in the observed features; and (3) $\textrm{corr} \in
\{0.1, 0.9\}$, where a larger value indicates a larger between-feature
correlation in the training set as compared to the test set.

To simulate covariate shift, we introduce the parameter
$\textrm{corr}$: the correlation of any covariate pair is $\pm
\hspace{0.5mm} \textrm{corr}$ in the training set, and $1 -
\textrm{corr}$ in the test set. We generate the responses as a linear
combination of the covariates. Moreover, we add Gaussian noise to the
covariates, mimicking settings where covariates are measured with
error. Lastly, we bin each covariate and response into $C = 10$
categories based on their quantiles. This results in a $10$-class
classification problem with ordinal covariates and responses.

For a fixed $n_c \in \{1,5\}$, $\textrm{noise} \in \{0, 0.5, 1.5\}$ and $\textrm{corr} \in \{0.1, 0.9\}$, our data generation process can be described as follows.

\begin{enumerate}
    \item Let $\textrm{train\_cov} = \textrm{corr} \cdot J_5 + (1 - \textrm{corr}) \cdot I_5$ and $\textrm{test\_cov} = (1 - \textrm{ corr}) \cdot J_5 + \textrm{corr} \cdot I_5$. Here, $J_5$ represents a $5 \times 5$ matrix whose entries are all 1, and $I_5$ represents a $5 \times 5$ identity matrix.
    \item Generate samples $\textrm{train\_x\_true}$ and $\textrm{test\_x\_true}$ from zero-mean multivariate normal distributions with covariance matrices $\textrm{train\_cov}$ and $\textrm{test\_cov}$, respectively. Each sample is of size 2,000.
    \item Introduce positively and negatively correlated covariates in the training samples by multiplying data in the first two features by $-1$. 
    \item Add Gaussian observation noises to the training and test samples. For the $n_c$ features which generate the response, add $0.4 \cdot \textrm{noise} \cdot \mathcal{N}(0,1)$; otherwise, add $0.3 \cdot \textrm{noise} \cdot \mathcal{N}(0,1)$. Let the resulting samples be $\textrm{train\_x}$ and $\textrm{test\_x}$.
    \item Generate the true coefficient for each of the $n_c$ features from $\mathcal{U}(0,10)$. \item Generate the training response $\textrm{train\_y}$, which is a linear combinations of the $n_c$ features of $\textrm{train\_x\_true}$ with the true coefficients as weights, plus a Gaussian noise from $\mathcal{N}(0,4)$. Generate the test response $\textrm{test\_y}$ in a similar manner.
    \item Bin each feature and response of $(\textrm{train\_x}, \textrm{train\_y})$ and $(\textrm{test\_x}, \textrm{test\_y})$ into 10 quantile-based categories. 
\end{enumerate}

\section{Implementation and hyperameter tuning process for competing models}
\label{app:hyp-tun}

We fit the proposed tabular extension of bidirectional attention model
to each training set, together with a few competing methods, namely
logistic regression (LR), random forests (RF), gradient boosting (GB)
and multilayer perceptron (MLP). We then evaluate the prediction
accuracy (Acc) and mean squared error (MSE) on the corresponding test
set. For each set of hyperparameters, we take the average of both
metrics across the 20 generated data sets.

We implement the proposed extension of bidirectional attention (ATN)
in Keras using a single-layer BERT \citep{devlin2018bert} with 5
heads, an embedding size of 20, and a feed-forward layer of dimension
5. We use the Adam optimizer with the default parameters, and a batch
and epoch size of 128 and 200, respectively. For the competing
methods, we use sklearn's implementation with hyperparameters chosen
via 5-fold cross-validation in classification accuracy.

For each data set, the hyperparameters of the random forest (RF),
gradient boosting (GB) and multilayer perceptron (MLP) models are
chosen via 5-fold cross-validation based on the classification
accuracy.

\textbf{Random forest.} We consider every combination of the following hyperparameters: (a) \textit{criterion}: \texttt{gini} or \texttt{entropy}; (b) \textit{n\_estimators}: 50, 100 or 200; and (c) \textit{max\_depth}: 1, 3 or \texttt{None}.

\textbf{Gradient boosting.} We consider every combination of the following hyperparameters: (a) \textit{learning\_rate}: 0.01, 0.1 or 1; (b) \textit{n\_estimators}: 50, 100 or 200; and (c) \textit{max\_depth}: 1, 3 or 5.

\textbf{Multilayer perceptron.} We consider every combination of the following hyperparameters: (a) \textit{hidden\_layer\_sizes}: (50,), (100,) or (100,50); (b) \textit{alpha}: 0.0001, 0.001 or 0.01; and (c) \textit{learning\_rate}: \texttt{constant} or \texttt{adaptive}.

\section{Details of the word analogy experiment}
\label{sec:analogy-details-expm}

\parhead{Data description.} We use the analogy data set first
introduced in \citet{pennington2014glove}. This data set contains
19,544 questions of the form ``\textit{a} is to \textit{b} as
\textit{c} is to ?", together with the correct answers. As an example,
the first question in the data set is ``Athens is to Greece as Baghdad
is to ?" (correct answer: Iraq). Overall, these questions can be
categorized into two groups: semantic (about people and places) and
syntactic (about word forms such as comparative, superlative and
plural). For each question, we look for the word $d \neq a,b,c$ in the
vocabulary such that the cosine similarity between $x_d$ and $x_b +
x_c - x_a$ is maximized; $x_i$ represents the embedding of word $i$.

We only include a question when all four words involved are present in
the vocabulary list of each model. Out of 19,544 questions in the data
set, 9,522 (49\%) of them satisfy this condition. Analyzing each
category separately, we find that the condition is satisfied for 2,278
(26\%) out of 8,869 semantic questions, and 7,244 (68\%) out of 10,675
syntactic questions.

\parhead{Models.} We consider three models: (1) BERT base uncased,
which is used in the original BERT paper \citep{devlin2018bert}; (2)
GloVe trained on Wikipedia \citep{pennington2014glove}; (3) word2vec
trained with CBOW \citep{mikolov2013distributed}. The embedding
dimensions of these models are 768, 300, 300 and 768, respectively,
while the vocabulary size are around 30K, 400K, 3M and 30K,
respectively. Since all questions in the data set consist of single
words (e.g., not \textit{golden\_retriever}). In order to perform a
fair comparison among these models, we only consider single words as
possible answers to each question; we also exclude non-words (e.g.,
\textit{[unused9]}, \textit{\#\# ?}) from the list of possible
answers. 
    
\section{Detailed analysis of embeddings for \gls{CBOW} and bidirectional attention}

\label{sec:detailed-analogy}

We begin with theoretically characterize under which conditions can
\gls{CBOW} embeddings exhibit linear word analogies. Adopting
\citeauthor{allen2019analogies}'s [\citeyear{allen2019analogies}]
argument for \gls{SGNS}, we extend the argument to both \gls{CBOW} and
attention-based token embeddings, thanks to the equivalence we
established in \Cref{prop:mlm-moe-equiv}.

\subsection{Linear word analogies in \gls{CBOW} embeddings}

To perform this theoretical analysis, we follow existing analyses about \gls{SGNS}: \citet{levy2014neural} showed that for a sufficiently large embedding dimension, embeddings from \gls{SGNS} satisfy
    $w_i^\top c_j = \log \left( \frac{p(w_i, c_j)}{p(w_i) p(c_j)}\right) - \log k = \textrm{PMI}(w_i, c_j) - \log k,$
where $k$ is the number of negative samples for each positive sample;
$W^{LOV}, C \in \mathbb{R}^{|V| \times p}$ are the center and context
embedding matrix, respectively. For each $i \in [|V|]$, $w_i^\top$
($c_i^\top$) is the $i$-th row of $W^{LOV}$ ($C$), which represents
the center (context) embedding of word $i$.

Using this result, \citet{allen2019analogies}
considered embeddings which factorize the unshifted PMI matrix, namely
$w_i^\top c_j = \textrm{PMI}(w_i, c_j)$, compactly written as $W^\top
C = \textrm{PMI}$. Through the ideas of \textit{paraphrases} and
\textit{word transformations}, they explained why linear relationships
exist for analogies on \gls{SGNS} word embeddings.

We next perform similar analyses for \gls{CBOW} and bidirectional
attention to characterize their conditions required for linear word
analogies.

\parhead{What matrix does CBOW (approximately) factorize?} \Cref{prop:supp-cbow-sim} is the CBOW version of \citeauthor{levy2014neural}'s [\citeyear{levy2014neural}] classical result on between-token similarities for \gls{SGNS}. The proof can be found in \Cref{sec:app-proof-simil-w2v}. 

\begin{prop}
\label{prop:supp-cbow-sim}
Consider CBOW without negative sampling. Using the same notation as before, we have
\begin{equation*}
    w_i^\top c_j \approx \log \left( \frac{p(w_i, c_j)}{p(c_j)} \right) + \log |V|.
\end{equation*}
\end{prop}

From \Cref{prop:supp-cbow-sim}, we know that CBOW approximately factorizes $M$, a $|V| \times |V|$ matrix such that 
$$M_{i,j} = \log \left(\frac{p(w_i, c_j)}{p(c_j)} \right) + \log |V|.$$

It is worth noting that this formula is similar to that for noise-contrastive estimation (NCE) as mentioned in \citet{levy2014neural}, with $\log |V|$ replaced by $- \log k$. Also, observe that $w_i^\top c_j > w_k^\top c_j$ if and only if $p(w_i, c_j) > p(w_k, c_j)$. 

We empirically verify \Cref{prop:supp-cbow-sim} using a toy corpus with a vocabulary size of 12. This corpus consists of 10,000 sentences, each of which has length 5. The corpus generation process is detailed in \Cref{sec:app-corp-gen-proc}. We then train a CBOW model with the whole sentence except the center word as the context. We choose the embedding dimension to be one of $\{30, 100, 300, 900\}$. For each dimension, we compute (1) the Spearman correlation between $w_i^\top c_j$ and $p(w_i, c_j) / p(c_j)$ for each $i,j$; and (2) the Pearson correlation between $w_i^\top c_j$ and $\log \left(p(w_i, c_j) / p(c_j) \right) + \log |V|$ for each $i,j$ such that the latter is well-defined. We obtain values of $(0.74, 0.77, 0.77, 0.77)$ for (1) and $(0.67, 0.71, 0.70, 0.71)$ for (2), which are reasonably high.

\parhead{The paraphrasing argument for CBOW.} We look at what it means for two word sets to paraphrase each other.

\begin{defn}[Definition D2 of \citet{allen2019analogies}]
\label{defn: paraph}
Let $\mathcal{E}$ be the set of all words in the vocabulary. Two word sets $\mathcal{W}, \mathcal{W}_* \subseteq \mathcal{E}$ are said to paraphrase each other if the paraphrase error $\rho^{\mathcal{W}, \mathcal{W}_*} \in \mathbb{R}^{|V|}$ is element-wise small, where
\begin{equation*}
    \rho^{\mathcal{W}, \mathcal{W}_*}_j = \log \left( \frac{p(c_j | \mathcal{W}_*)}{p(c_j | \mathcal{W})} \right)
\end{equation*}
for every $c_j \in \mathcal{E}$.
\end{defn}
Intuitively, ``word sets paraphrase one another if they induce equivalent distributions over context words". When $\mathcal{W}$ and $\mathcal{W}_*$ paraphrase each other, we write $\mathcal{W} \approx_P \mathcal{W}_*$. From \Cref{defn: paraph}, we observe that $\mathcal{W} \approx_P \mathcal{W}_*$ if and only if $\mathcal{W}_* \approx_P \mathcal{W}$. Also, we implicitly require both $p(\mathcal{W}_*)$ and $p(\mathcal{W})$ to be positive. This is exactly Assumption A3 in the original paper. We now provide an equivalent version of their Lemma 2 for the matrix $M$. Here, $M_i^\top$ denotes the $i$-th row of $M$. The proof is provided in \Cref{sec:pf-lemma-5}.
\begin{lemma}
\label{lemma:supp-lem-2-for-m}
For any word sets $\mathcal{W}, \mathcal{W}_* \subseteq \mathcal{E}$ with the same cardinality, we have
\begin{equation*}
    \begin{split}
        \sum_{w_i \in \mathcal{W}_*} M_i &= \sum_{w_i \in \mathcal{W}} M_i + \rho^{\mathcal{W}, \mathcal{W}_*} + \sigma^{\mathcal{W}} - \sigma^{\mathcal{W}_*} + \delta^{\mathcal{W}, \mathcal{W}_*} \\
        &= \sum_{w_i \in \mathcal{W}} M_i + \xi^{\mathcal{W}, \mathcal{W}_*} + \sigma^{\mathcal{W}} - \sigma^{\mathcal{W}_*},
    \end{split}
\end{equation*}
where 
$$\sigma_j^{\mathcal{W}} = \log \left( \frac{p(\mathcal{W} | c_j)}{\prod_{w_i \in \mathcal{W}} p(w_i|c_j)} \right),$$
$$\sigma_j^{\mathcal{W}_*} = \log \left( \frac{p(\mathcal{W}_* | c_j)}{\prod_{w_i \in \mathcal{W}_*} p(w_i|c_j)} \right),$$

$\delta_j^{\mathcal{W}, \mathcal{W}_*}  = \log \left( \frac{p(\mathcal{W}_*)}{p(\mathcal{W})} \right)$, and $\xi_j^{\mathcal{W}, \mathcal{W}_*} = \log \left( \frac{p( \mathcal{W}_* | c_j)}{p( \mathcal{W} | c_j)} \right)$.

\end{lemma}
 
\Cref{prop:supp-anal-w2v}, which is equivalent to Corollary 2.3 of \citet{allen2019analogies}, follows from multiplying both sides of the equations in \Cref{lemma:supp-lem-2-for-m} by $C^{\dagger} = (C C^\top)^{-1} C$ (assuming $C$ has full row rank) and setting $\mathcal{W} =  \{ w_b, w_{a^*}\}$ and $\mathcal{W}_* = \{ w_{b^*}, w_a \}$.
\begin{prop}
\label{prop:supp-anal-w2v}
    Given any $w_a, w_{a^*}, w_b, w_{b^*} \in \mathcal{E}$, we have
    \begin{align*}
            w_{b^*} 
            &= w_{a^*} - w_a + w_b + C^{\dagger}(\rho^{\mathcal{W}, \mathcal{W}_*} + \sigma^{\mathcal{W}} - \sigma^{\mathcal{W}_*} + \delta^{\mathcal{W}, \mathcal{W}_*}) \\
            &= w_{a^*} - w_a + w_b + C^{\dagger}(\xi^{\mathcal{W}, \mathcal{W}_*} + \sigma^{\mathcal{W}} - \sigma^{\mathcal{W}_*}),
    \end{align*}
    where $\mathcal{W} = \{ w_b, w_{a^*}\}$ and $\mathcal{W}_* = \{ w_{b^*}, w_a \}$.
\end{prop}
From \Cref{prop:supp-anal-w2v}, we see that when $\mathcal{W} \approx_P \mathcal{W}_*$, and $\sigma^{\mathcal{W}}$, $\sigma^{\mathcal{W}_*}$ and $\delta^{\mathcal{W}, \mathcal{W}_*}$ are small, we have $w_{b^*} \approx w_{a^*} - w_a + w_b$. By definition, $\sigma^{\mathcal{W}}$ ($\sigma^{\mathcal{W}_*}$) is small when all $w_i \in \mathcal{W}$ ($w_i \in \mathcal{W}_*$) are approximately conditionally independent given $c_j$, and $\delta^{\mathcal{W}, \mathcal{W}_*}$ is small when $p(\mathcal{W}) \approx p(\mathcal{W}_*)$. Following the connection between analogies and word transformations described in Sections 6.3 and 6.4 of \citet{allen2019analogies}, we now have an approximately linear relationship for CBOW embeddings with some error terms mentioned above.

Alternatively, we can modify \Cref{defn: paraph} so that $\mathcal{W} \approx_P \mathcal{W_*}$ if and only if $\xi^{\mathcal{W}, \mathcal{W}_*}$ (instead of $\rho^{\mathcal{W}, \mathcal{W}_*}$) is element-wise small. Now, our error terms only depend on the approximate conditional independence of $w_i$'s given $c_j$.

\parhead{Does this linear relationship also hold for context embeddings?} In other words, if $w_r + w_s \approx w_t + w_u$, do we have $c_r + c_s \approx c_t + c_u$? \Cref{prop:supp-anal-ctx}, whose proof is provided in \Cref{sec:app-proof-w2v-anal-context}, answers the question.

\begin{prop}
\label{prop:supp-anal-ctx}
    Let $\mathcal{W} = \{r,s\}$ and $\mathcal{W}_* = \{t,u\}$. Assume $p(\mathcal{W}) \approx p(\mathcal{W_*})$ and $w_i \in \mathcal{W}$ ($w_i \in \mathcal{W_*}$) are approximately marginally independent. Also, assume that $W$ has full row rank. If $w_r + w_s \approx w_t + w_u$, then $c_r + c_s \approx c_t + c_u$.
\end{prop}

So far, we have argued that both the center and context embeddings of CBOW exhibit linear structures under some assumptions. We now extend this argument to MLM with self-attention, and show that the same conclusion holds under stronger assumptions.

\subsection{Linear word analogies in attention-based embeddings}

Similar to \Cref{sec:lin-rel-cbow}, we compute the matrix MLM with self-attention factorized and construct a paraphrasing argument to show linear structures in the learned embeddings.

\parhead{What matrix does MLM with self-attention (approximately) factorize?} To make calculations tractable, we exclude both residual connections and positional encodings. Let the masked sentence be $(a_1, \cdots, a_S)$. As before, let $m \in [S]$ and $b \in [|V|]$ denote the masked position and masked word, respectively. This means $a_i \in [|V|]$ for every $i \neq m$ and $a_m = |V| + 1$. From \Cref{prop:mlm-obj}, the loss for this instance is given by
\begin{equation}
\label{eq:sim-attn-loss}
\begin{split}
    -\sum_{j=1}^S \frac{\tau_{a_j}}{\sum_{j=1}^S \tau_{a_j}} w_b^\top c_{a_j} + \log\left( \sum_{k=1}^{|V|} \exp\left( \sum_{j=1}^S \frac{\tau_{a_j}}{\sum_{j=1}^S \tau_{a_j}} w_k^\top c_{a_j}  \right)\right),
\end{split}
\end{equation}
where $\tau_j = \exp\left( \frac{c_j ^\top W^{KQ} c_{|V|+1}}{\sqrt{d_w}} \right)$. \Cref{prop:supp-sim-attn} approximates the matrix factorized by the attention objective, given all $\tau_j$ values for each $j \in [|V| + 1]$. The proof is similar to that of \Cref{prop:supp-cbow-sim}, and therefore omitted.

\begin{prop}
\label{prop:supp-sim-attn}
    Consider the attention objective as in \Cref{eq:sim-attn-loss}. We have
    \begin{equation}
    \label{eq:sim-attn}
    \begin{split}
         w_i^\top c_j 
         &\approx \frac{|V| \sum_{(i,j)} \gamma_{j}^i - \left(\sum_{(1,j)} \gamma_{j}^1 + \cdots + \sum_{(|V|,j)} \gamma_{j}^{|V|} \right)}{S \left(\sum_{(1,j)} (\gamma_{j}^1)^2 + \cdots + \sum_{(|V|,j)} (\gamma_{j}^{|V|})^2 \right)},
    \end{split}
    \end{equation}
where for a center-context pair $(d,j)$ in the masked sentence $(a_1, \cdots, a_S)$, we define $\gamma_j^d = \tau_j / \sum_{s=1}^S \tau_{a_s}$.
\end{prop}
In other words, MLM with self-attention approximately factorizes a $|V| \times |V|$ matrix whose $(i,j)$-th entry is given by \Cref{eq:sim-attn}. It is important to note that unlike in CBOW, the token embedding for each word $i$ is $c_i$ (the \textit{context} embedding), and not $w_i$ (the \textit{center} embedding). In the case where $\tau_j$ is approximately the same for every $j \in [|V|+1]$, our problem approximately reduces to a vanilla CBOW. In particular, we always have $\gamma_j^d \approx 1/S$, whence \Cref{prop:supp-sim-attn} yields $w_i^\top c_j \approx \frac{p(w_i, c_j)}{p(c_j)} \cdot |V| - 1 \approx \log \left( \frac{p(w_i, c_j)}{p(c_j)} \right) + \log |V|$. Using \Cref{prop:supp-anal-w2v}, we argue that the resulting embeddings approximately form a linear relationship, up to some error terms. 

\parhead{The paraphrasing argument for MLM with self-attention.} We first define
\begin{equation*}
    \tilde{c}_j := \frac{S \left(\sum_{(1,j)} (\gamma_{j}^1)^2 + \cdots + \sum_{(|V|,j)} (\gamma_{j}^{|V|})^2 \right)}{\sum_{(1,j)} \gamma_{j}^1 + \cdots + \sum_{(|V|,j)} \gamma_{j}^{|V|} } c_j
\end{equation*}
for every $j \in [|V|+1]$. This means
\begin{equation*}
\begin{split}
     w_i^\top \tilde{c}_j 
     &\approx \frac{|V| \sum_{(i,j)} \gamma_{j}^i}{\sum_{(1,j)} \gamma_{j}^1 + \cdots + \sum_{(|V|,j)} \gamma_{j}^{|V|} } - 1  \\
     &\approx \log \left( \frac{ \sum_{(i,j)} \gamma_{j}^i}{\sum_{(1,j)} \gamma_{j}^1 + \cdots + \sum_{(|V|,j)} \gamma_{j}^{|V|} }\right) + \log |V|,
\end{split}
\end{equation*}
where we used the approximation $x \approx \log(1+x)$. Previously, $p(w_i, c_j)$ represents a population quantity which is estimated by $\#(w_i,c_j)/D$, where $D$ is a normalizing constant, and $p(c_j) = \sum_i p(w_i, c_j)$. We now define $\bar{p}(w_i, c_j)$, a population quantity which is estimated by $\sum_{(i,j)} \gamma_j^i/E$ for some normalizing constant $E$. We have
\begin{equation*}
\label{eq:sim-attn-transf}
    w_i^\top \tilde{c}_j \approx \log \left( \frac{\bar{p}(w_i, c_j)}{\bar{p}(c_j)} \right) + \log |V|,
\end{equation*}
where $\bar{p}(c_j) = \sum_i \bar{p}(w_i,c_j)$. Note that unlike $p$, $\bar{p}$ is not symmetric, i.e., $\bar{p}(w_i, c_j) \neq \bar{p}(w_j,c_i)$. Having defined $\bar{p}$, we are ready to state \Cref{lemma:supp-lem-5-for-n}, which is a version of \Cref{lemma:supp-lem-2-for-m} for the matrix $N$, where $$N_{i,j} = \log \left( \frac{\bar{p}(w_i, c_j)}{\bar{p}(c_j)} \right) + \log |V|.$$ 
Here, $N_i^\top$ denotes the $i$-th row of $N$. The proof is analogous to that of \Cref{lemma:supp-lem-2-for-m} and is thus omitted.

\begin{lemma}
\label{lemma:supp-lem-5-for-n}
For any word sets $\mathcal{W}, \mathcal{W}_* \subseteq \mathcal{E}$ with the same cardinality, we have
\begin{equation*}
\begin{split}
    \sum_{w_i \in \mathcal{W}_*} N_i &= \sum_{w_i \in \mathcal{W}} N_i + \bar{\rho}^{\mathcal{W}, \mathcal{W}_*} + \bar{\sigma}^{\mathcal{W}} - \bar{\sigma}^{\mathcal{W}_*} + \bar{\delta}^{\mathcal{W}, \mathcal{W}_*} \\
    &= \sum_{w_i \in \mathcal{W}} N_i + \bar{\xi}^{\mathcal{W}, \mathcal{W}_*} + \bar{\sigma}^{\mathcal{W}} - \bar{\sigma}^{\mathcal{W}_*},
\end{split}
\end{equation*}
where 
$$\bar{\sigma}_j^{\mathcal{W}} = \log \left( \frac{\bar{p}(\mathcal{W} | c_j)}{\prod_{w_i \in \mathcal{W}} \bar{p}(w_i|c_j)} \right),$$
$$\bar{\sigma}_j^{\mathcal{W}_*} = \log \left( \frac{\bar{p}(\mathcal{W}_* | c_j)}{\prod_{w_i \in \mathcal{W}_*} \bar{p}(w_i|c_j)} \right),$$ 
$\bar{\rho}^{\mathcal{W}, \mathcal{W}_*}_j = \log \left( \frac{\bar{p}(c_j | \mathcal{W}_*)}{\bar{p}(c_j | \mathcal{W})} \right)$ , $\bar{\delta}_j^{\mathcal{W}, \mathcal{W}_*}  = \log \left( \frac{\bar{p}(\mathcal{W}_*)}{\bar{p}(\mathcal{W})} \right)$, and $\bar{\xi}_j^{\mathcal{W}, \mathcal{W}_*} = \log \left( \frac{\bar{p}( \mathcal{W}_* | c_j)}{\bar{p}( \mathcal{W} | c_j)} \right)$.
\end{lemma}
\Cref{prop:supp-anal-w2v-attn,prop:supp-anal-ctx-attn} are the attention versions of \Cref{prop:supp-anal-w2v,prop:supp-anal-ctx}. The proof of \Cref{prop:supp-anal-w2v-attn} follows from multiplying both sides of the equations in \Cref{lemma:supp-lem-5-for-n} by $\tilde{C}^{\dagger} = (\tilde{C} \tilde{C}^\top)^{-1} C$ (assuming $\tilde{C}$ has full row rank) and setting $\mathcal{W} =  \{ w_b, w_{a^*}\}$ and $\mathcal{W}_* = \{ w_{b^*}, w_a \}$.
The proof of \Cref{prop:supp-anal-ctx-attn} can be found in \Cref{sec:app-proof-attn-anal-context}. 
\begin{prop}
\label{prop:supp-anal-w2v-attn}
    Given any $w_a, w_{a^*}, w_b, w_{b^*} \in \mathcal{E}$, we have
    \begin{align*}
            w_{b^*} 
            &= w_{a^*} - w_a + w_b + \tilde{C}^{\dagger}(\bar{\rho}^{\mathcal{W}, \mathcal{W}_*} + \bar{\sigma}^{\mathcal{W}} - \sigma^{\mathcal{W}_*} + \bar{\delta}^{\mathcal{W}, \mathcal{W}_*}) \\
            &= w_{a^*} - w_a + w_b + \tilde{C}^{\dagger}(\bar{\xi}^{\mathcal{W}, \mathcal{W}_*} + \bar{\sigma}^{\mathcal{W}} - \bar{\sigma}^{\mathcal{W}_*}),
    \end{align*}
    where $\mathcal{W} = \{ w_b, w_{a^*}\}$ and $\mathcal{W}_* = \{ w_{b^*}, w_a \}$.
\end{prop}

\begin{prop}
\label{prop:supp-anal-ctx-attn}
    Let $\mathcal{W} = \{r,s\}$ and $\mathcal{W}_* = \{t,u\}$. Assume $\bar{p}(\mathcal{W}) \approx \bar{p}(\mathcal{W_*})$ and $w_i \in \mathcal{W}$ ($w_i \in \mathcal{W_*}$) are approximately marginally independent. Also, assume that $W$ has full row rank and $\bar{p}(w_i, c_j) \approx \bar{p}(w_j,c_i)$. If $w_r + w_s \approx w_t + w_u$, then $\tilde{c}_r + \tilde{c}_s \approx \tilde{c}_t + \tilde{c}_u$.
\end{prop}

\parhead{What do we learn from these results?} One important takeaway is that the sufficient conditions to obtain linear relationships are stronger in the case of MLM with self-attention as compared to CBOW. 
Concretely, we need $\bar{p}$ to be approximately symmetric. Even when this is satisfied, the linear relationships hold for the transformed embeddings $\tilde{c}_i$'s instead of the token embeddings $c_i$'s. Under an additional assumption that
\begin{equation*}
   \zeta_j := \frac{\sum_{(1,j)} (\gamma_{j}^1)^2 + \cdots + \sum_{(|V|,j)} (\gamma_{j}^{|V|})^2 }{\sum_{(1,j)} \gamma_{j}^1 + \cdots + \sum_{(|V|,j)} \gamma_{j}^{|V|} }
\end{equation*}
is approximately the same for each $j$ (e.g., when $\tau_j$ is approximately the same for every $j$), we approximately have linear relationships for the token embeddings $c_i$'s.

\textit{\textbf{Remarks.} It is easy to see that our result can technically be extended to incorporate positional encodings by considering each (word, position) pair as a unit. In particular, analogies are drawn between (word, position) units.}

\section{Proof of Proposition \ref{prop:supp-cbow-sim}}
\label{sec:app-proof-simil-w2v}

\textbf{Proposition \ref{prop:supp-cbow-sim}}. \textit{Consider CBOW without negative sampling. Using the same notation as before, we have
\begin{equation*}
    w_i^\top c_j \approx \log \left( \frac{p(w_i, c_j)}{p(c_j)} \right) + \log |V|.
\end{equation*}}

\textit{Proof.} For simplicity, we assume that the window size is always $2m$. Consider an instance with $i$ as the center word and $j \in J$ as the context words. The loss for this instance can be approximated as
\begin{equation*}
    \begin{split}
        &-\frac{\sum_{j \in J} w_i^\top c_j}{2m} + \log \left( \sum_{k=1}^{|V|} \exp \left( \frac{\sum_{j \in J} w_k^\top c_j}{2m}\right) \right) \\
        &\approx -\frac{\sum_{j \in J} w_i^\top c_j}{2m} + \log \left( \sum_{k=1}^{|V|} \left( 1 + \frac{\sum_{j \in J} w_k^\top c_j}{2m} + \frac{(\sum_{j \in J} w_k^\top c_j)^2}{8m^2} \right) \right)  \\
        &= -\frac{\sum_{j \in J} w_i^\top c_j}{2m} + \log   |V| + \log \left( 1 + \frac{\sum_{k=1}^{|V|} \left( \sum_{j \in J} w_k^\top c_j\right)}{2m |V|} + \frac{\sum_{k=1}^{|V|} \left( \sum_{j \in J} w_k^\top c_j\right)^2}{8m^2 |V|} \right)  \\
        &\approx -\frac{\sum_{j \in J} w_i^\top c_j}{2m} + \log |V| + \frac{\sum_{k=1}^{|V|} \left( \sum_{j \in J} w_k^\top c_j\right)}{2m |V|} + \frac{\sum_{k=1}^{|V|} \left( \sum_{j \in J} w_k^\top c_j\right)^2}{8m^2 |V|} \\
        &\leq -\frac{\sum_{j \in J} w_i^\top c_j}{2m} + \log |V| + \frac{\sum_{k=1}^{|V|} \left( \sum_{j \in J} w_k^\top c_j\right)}{2m |V|} + \frac{\sum_{k=1}^{|V|} \left( \sum_{j \in J} (w_k^\top c_j)^2\right)}{4m |V|},
    \end{split}
\end{equation*}
where we used the Taylor expansions $\exp(x) \approx 1 + x + x^2/2$ and $\log(1 + x) \approx x$, as well as the Cauchy-Schwarz inequality. Ignoring the constant $\log |V|$ and multiplying by $2m|V|$, the approximate loss can be written as
\begin{equation*}
    - |V| \sum_{j \in J} w_i^\top c_j + \sum_{k=1}^{|V|} \left( \sum_{j \in J} w_k^\top c_j\right) + \frac{1}{2} \sum_{k=1}^{|V|} \left( \sum_{j \in J} (w_k^\top c_j)^2\right).
\end{equation*}
Summing this over all instances and only extracting terms which depend on $w_i^\top c_j$, we have the following loss which we want to minimize:
\begin{equation*}
    \ell(i, j) = - |V| \cdot \#(w_i, c_j) w_i^\top c_j + \#(c_j) w_i^\top c_j + \frac{1}{2} \#(c_j) (w_i^\top c_j)^2.
\end{equation*}
Taking derivative with respect to $w_i^\top c_j$ and setting it to 0 yields 
\begin{equation*}
    w_i^\top c_j = \left(\frac{\#(w_i,c_j)}{\#(c_j)}\right) \cdot |V| - 1 = \left(\frac{p(w_i, c_j)}{p(c_j)} \cdot |V| \right) - 1.
    \end{equation*}
The approximation $x \approx \log(1+x)$ completes the proof.

\section{Corpus generation process}
\label{sec:app-corp-gen-proc}
\begin{enumerate}
    \item Consider four subjects (mathematics, statistics, sociology and history) and four adjectives (fun, boring, easy and difficult). Assign scores to each subject which represents the level of each adjective:
    \begin{enumerate}
        \item mathematics: (4, 2, 4, 2).
        \item statistics: (6, 0, 5, 1).
        \item sociology: (1, 5, 2, 4).
        \item history: (0, 6, 0, 6).
    \end{enumerate}
    \item Consider three types of sentence:
        \begin{enumerate}
            \item Type 1: I like \texttt{subj1} and \texttt{subj2}, where \texttt{subj1} and \texttt{subj2} are independently chosen from the list of subjects with probability $(4/11, 5/11, 1/11, 1/11)$. 
            \item Type 2: \texttt{subj1} and \texttt{subj2} is \texttt{adj}, where \texttt{subj1} and \texttt{subj2} are independently chosen from the list of subjects with uniform probability, and  \texttt{adj} is chosen from the list of adjectives with probability proportional to the sum of the scores of \texttt{subj1} and \texttt{subj2}.
            \item Type 3: \texttt{subj} is \texttt{adj1} and \texttt{adj2}, where \texttt{subj} is chosen from the list of subjects with uniform probability, and \texttt{adj1} and \texttt{adj2} are independently chosen from the list of adjectives with probability proportional to the score of \texttt{subj}.
        \end{enumerate}
    \item To generate each sentence, we first randomly choose the sentence type with uniform probability. We then form the sentence following the process above.
\end{enumerate}

\section{Proof of Lemma \ref{lemma:supp-lem-2-for-m}}
\label{sec:pf-lemma-5}
\textbf{Lemma \ref{lemma:supp-lem-2-for-m}}. \textit{For any word sets $\mathcal{W}, \mathcal{W}_* \subseteq \mathcal{E}$ with the same cardinality, we have
\begin{equation*}
    \begin{split}
        \sum_{w_i \in \mathcal{W}_*} M_i &= \sum_{w_i \in \mathcal{W}} M_i + \rho^{\mathcal{W}, \mathcal{W}_*} + \sigma^{\mathcal{W}} - \sigma^{\mathcal{W}_*} + \delta^{\mathcal{W}, \mathcal{W}_*} \\
        &= \sum_{w_i \in \mathcal{W}} M_i + \xi^{\mathcal{W}, \mathcal{W}_*} + \sigma^{\mathcal{W}} - \sigma^{\mathcal{W}_*},
    \end{split}
\end{equation*}
where $\sigma_j^{\mathcal{W}} = \log \left( \frac{p(\mathcal{W} | c_j)}{\prod_{w_i \in \mathcal{W}} p(w_i|c_j)} \right)$, $\sigma_j^{\mathcal{W}_*} = \log \left( \frac{p(\mathcal{W}_* | c_j)}{\prod_{w_i \in \mathcal{W}_*} p(w_i|c_j)} \right)$, $\delta_j^{\mathcal{W}, \mathcal{W}_*}  = \log \left( \frac{p(\mathcal{W}_*)}{p(\mathcal{W})} \right)$, and $\xi_j^{\mathcal{W}, \mathcal{W}_*} = \log \left( \frac{p( \mathcal{W}_* | c_j)}{p( \mathcal{W} | c_j)} \right)$.}

\textit{Proof.} Observe that $p(c_j | \mathcal{W}_*) = \frac{p(\mathcal{W}_* | c_j) p(c_j)}{p(\mathcal{W}_*)}$ and $p(c_j | \mathcal{W}) = \frac{p(\mathcal{W} | c_j) p(c_j)}{p(\mathcal{W})}$, whence $\rho_j^{\mathcal{W}, \mathcal{W}_*} = \log \left( \frac{p(c_j | \mathcal{W}_*)}{p(c_j | \mathcal{W})} \right) = \log \left( \frac{p( \mathcal{W}_* | c_j) }{p( \mathcal{W} | c_j) } \right) + \log \left( \frac{p(\mathcal{W})}{p(\mathcal{W}_*)} \right)$. We have
\begin{align*}
    \sum_{w_i \in \mathcal{W}_*} M_i - \sum_{w_i \in \mathcal{W}} M_i
    &=\sum_{w_i \in \mathcal{W}_*} \log \left(\frac{p(w_i, c_j)}{p(c_j)} \right) - \sum_{w_i \in \mathcal{W}} \log \left(\frac{p(w_i, c_j)}{p(c_j)} \right) \\
    &= \log \prod_{w_i \in \mathcal{W}_*} p(w_i | c_j) - \log \prod_{w_i \in \mathcal{W}} p(w_i | c_j) \\
    \begin{split}
    &= \log \left( \frac{ \prod_{w_i \in \mathcal{W}_*} p(w_i | c_j)}{ \prod_{w_i \in \mathcal{W}} p(w_i | c_j)} \right) + \log \left( \frac{p(\mathcal{W}_*)}{p(\mathcal{W}_*)} \right) + \log \left( \frac{p(\mathcal{W})}{p(\mathcal{W})} \right) \\
    &\qquad + \log \left( \frac{p(\mathcal{W}_* | c_j)}{p(\mathcal{W}_* | c_j)} \right) + \log \left( \frac{p(\mathcal{W} | c_j)}{p(\mathcal{W} | c_j)} \right)
    \end{split}
    \\
    \begin{split}
    &= \log \left( \frac{p( \mathcal{W}_* | c_j) }{p( \mathcal{W} | c_j) } \right) + \log \left( \frac{p(\mathcal{W})}{p(\mathcal{W}_*)} \right) + \log \left( \frac{p(\mathcal{W} | c_j)}{\prod_{w_i \in \mathcal{W}} p(w_i|c_j)} \right) \\
    &\qquad - \log \left( \frac{p(\mathcal{W}_* | c_j)}{\prod_{w_i \in \mathcal{W}_*} p(w_i|c_j)} \right) + \log \left( \frac{p(\mathcal{W}_*)}{p(\mathcal{W})} \right)
    \end{split}
    \\
    &= \rho_j^{\mathcal{W}, \mathcal{W}_*} + \sigma_j^{\mathcal{W}} - \sigma_j^{\mathcal{W}_*} + \delta_j^{\mathcal{W}, \mathcal{W}_*}.
\end{align*}
Also, 
\begin{align*}
    \rho_j^{\mathcal{W}, \mathcal{W}_*} + \delta_j^{\mathcal{W}, \mathcal{W}_*} &= \log \left( \frac{p( \mathcal{W}_* | c_j) }{p( \mathcal{W} | c_j) } \right) + \log \left( \frac{p(\mathcal{W})}{p(\mathcal{W}_*)} \right) + \log \left( \frac{p(\mathcal{W}_*)}{p(\mathcal{W})} \right) \\
    &= \xi_j^{\mathcal{W}, \mathcal{W}_*},
\end{align*}
which completes the proof.

\section{Proof of Proposition \ref{prop:supp-anal-ctx}}
\label{sec:app-proof-w2v-anal-context}
\textbf{Proposition \ref{prop:supp-anal-ctx}}. \textit{
    Let $\mathcal{W} = \{r,s\}$ and $\mathcal{W}_* = \{t,u\}$. Assume $p(\mathcal{W}) \approx p(\mathcal{W_*})$ and $w_i \in \mathcal{W}$ ($w_i \in \mathcal{W_*}$) are approximately marginally independent. Also, assume that $W$ has full row rank. If $w_r + w_s \approx w_t + w_u$, then $c_r + c_s \approx c_t + c_u$.
}

\textit{Proof.} For any $c_v \in \mathcal{E}$, we have $(w_r + w_s)^\top c_v \approx (w_t + w_u)^\top c_v$. From \Cref{prop:cbow-sim}, this expression can be simplified as $\log p(w_r, c_v) + \log p(w_s, c_v) \approx \log p(w_t, c_v) + \log p(w_u, c_v)$. This implies $\log p(w_v, c_r) + \log p(w_v, c_s) \approx \log p(w_v, c_t) + \log p(w_v, c_u)$. Observe that
    \begin{align*}
            &w_v^\top (c_r + c_s - c_t - c_u) \\ 
            &= (\log p(w_v, c_r) + \log p(w_v, c_s) - \log p(w_v, c_t) - \log p(w_v, c_u)) + \log \left( \frac{p(c_t) p(c_u)}{p(c_r) p(c_s)} \right) \\
            &\approx 0 + \log \left( \frac{p(\mathcal{W}_*)}{p(\mathcal{W})} \right) \\
            &\approx 0.
    \end{align*}
Since this holds for every $v$ and $W$ has full row rank, we conclude that $c_r + c_s \approx c_t + c_u$, completing the proof.

\section{Proof of Proposition \ref{prop:supp-anal-w2v-attn}}
\label{sec:app-proof-attn-anal-context}
\textbf{Proposition \ref{prop:supp-anal-w2v-attn}.} Let $\mathcal{W} = \{r,s\}$ and $\mathcal{W}_* = \{t,u\}$. Assume $\bar{p}(\mathcal{W}) \approx \bar{p}(\mathcal{W_*})$ and $w_i \in \mathcal{W}$ ($w_i \in \mathcal{W_*}$) are approximately marginally independent. Also, assume that $W$ has full row rank and $\bar{p}(w_i, c_j) \approx \bar{p}(w_j,c_i)$. If $w_r + w_s \approx w_t + w_u$, then $\tilde{c}_r + \tilde{c}_s \approx \tilde{c}_t + \tilde{c}_u$.

\textit{Proof.} For any $\tilde{c}_v \in \mathcal{E}$, we have $(w_r + w_s)^\top \tilde{c}_v = (w_t + w_u)^\top \tilde{c}_v$. From \Cref{eq:sim-attn-transf}, this expression can be simplified as $\log \bar{p}(w_r, c_v) + \log \bar{p}(w_s, c_v) \approx \log \bar{p}(w_t, c_v) + \log \bar{p}(w_u, c_v)$. By the assumption that $\bar{p}(w_i, c_j) \approx \bar{p}(w_j,c_i)$, this implies $\log \bar{p}(w_v, c_r) + \log \bar{p}(w_v, c_s) \approx \log \bar{p}(w_v, c_t) + \log \bar{p}(w_v, c_u)$. Observe that
    \begin{align*}
            &w_v^\top (\tilde{c}_r + \tilde{c}_s - \tilde{c}_t - \tilde{c}_u) \\
            &= (\log \bar{p}(w_v, c_r) + \log \bar{p}(w_v, c_s) - \log \bar{p}(w_v, c_t) - \log \bar{p}(w_v, c_u)) + \log \left( \frac{\bar{p}(c_t) \bar{p}(c_u)}{\bar{p}(c_r) \bar{p}(c_s)} \right) \\
            &\approx 0 + \log \left( \frac{\bar{p}(\mathcal{W}_*)}{\bar{p}(\mathcal{W})} \right) \\
            &\approx 0.
    \end{align*}
Since this holds for every $v$ and $W$ has full row rank, we conclude that $\tilde{c}_r + \tilde{c}_s \approx \tilde{c}_t + \tilde{c}_u$, completing the proof.

\end{document}